\pdfoutput=1

\documentclass[11pt]{article}

\usepackage{ACL2023}

\usepackage{times}
\usepackage{latexsym}

\usepackage[T1]{fontenc}

\usepackage[utf8]{inputenc}

\usepackage{microtype}

\usepackage{hyperref}       
\usepackage{url}            
\usepackage{booktabs}       
\usepackage{amsfonts}       

\usepackage{graphicx}
\usepackage{listings}
\usepackage[normalem]{ulem}
\usepackage{multirow}
\usepackage{array}

\def\bi{\begin{itemize}}
	
	\def\ei{\end{itemize}}
\def\beq{\begin{equation}}
	\def\eeq#1{\label{#1}\end{equation}}
\def\ba{\begin{array}}
	\def\ea{\end{array}}

\newcommand{\cblu}{\color{blue}}
\newcommand{\cred}{\color{red}}

\long\def\BOC#1\EOC{\message{(Commented text )}}
\long\def\BOCC#1\EOCC{\message{(Commented text )}}
\long\def\BOCCC#1\EOCCC{\message{(Commented text )}}

\long\def\NBB#1{}

\long\def\NOWRITE#1{}

\title{Coupling Large Language Models with Logic Programming \\ for Robust and General Reasoning from Text} 

\author{Zhun Yang, Adam Ishay $^1$\\
  $^1$ Arizona State University \\
  \texttt{\{zyang90,aishay\}@asu.edu} \\\And
  Joohyung Lee $^{1,2}$\\
  $^2$ Samsung Research \\
  \texttt{joolee@asu.edu} \\}

\begin{document}
\maketitle
\begin{abstract}
While large language models (LLMs), such as GPT-3, appear to be robust and general, their reasoning ability is not at a level to compete with the best models trained for specific natural language reasoning problems. In this study, we observe that a large language model can serve as a highly effective few-shot semantic parser. It can convert natural language sentences into a logical form that serves as input for answer set programs, a logic-based declarative knowledge representation formalism. The combination results in a robust and general system that can handle multiple question-answering tasks without requiring retraining for each new task. It only needs a few examples to guide the LLM's adaptation to a specific task, along with reusable ASP knowledge modules that can be applied to multiple tasks. We demonstrate that this method achieves state-of-the-art performance on several NLP benchmarks, including bAbI, StepGame, CLUTRR, and gSCAN. Additionally, it successfully tackles robot planning tasks that an LLM alone fails to solve. 
\end{abstract}

\section{Introduction}  \label{sec:intro}

A typical way to handle a question-answering task is to train a neural network model on large training data and test it on similar data. Such models work well with linguistic variability and ambiguity but often learn statistical features and correlations rather than true reasoning \cite{ruder21benchmarking}, which makes them not robust, lack generalization, and difficult to interpret. 

Alternatively, transformer-based large language models (LLMs) have recently shown wide success on many downstream tasks, demonstrating general reasoning capability on diverse tasks without being retrained.
However, when we restrict our attention to individual NLP reasoning benchmarks, they usually do not perform as well as state-of-the-art models despite various efforts to improve accuracy through prompt engineering \cite{wei22chain,zhou22least}.

Similarly, LLMs gained attention for plan generation for robots due to the rich semantic knowledge they acquired about the world \cite{ahn22can,huang22inner,zeng22socratic}.
However, LLMs are known to perform shallow reasoning and cannot find complex plans \cite{valmeekam22large}.

In another context, \citet{nye21improving} note that while LLMs are good at System-1 thinking, their outputs are often inconsistent and incoherent.
This is because LLMs are trained to predict subsequent words in a sequence and do not appear to have a deep understanding of concepts such as cause and effect, logic, and probability, which are important for reasoning.

Nevertheless, we note that the rich semantic knowledge that LLMs possess makes them effective general-purpose few-shot semantic parsers that can convert linguistically variable natural language sentences into atomic facts that serve as input to logic programs. We also note that the fully declarative nature of answer set programs \cite{lif08,bre11} makes them a good pair with the LLM semantic parsers, providing interpretable and explainable reasoning on the parsed result of the LLMs using background knowledge. 
Combining large language models and answer set programs leads to an attractive dual-process, neuro-symbolic reasoning that works across multiple QA tasks without retraining for individual tasks. 

We tested this idea with several NLP benchmarks, bAbI \cite{weston15towards}, StepGame \cite{shi22stepgame},  CLUTRR \cite{sinha19clutrr}, and gSCAN \cite{ruis20benchmark}, by applying the same dual-system model and achieved state-of-the-art performance in all of them. 
Furthermore, the high accuracy and transparency allow us to easily identify the source of errors, making our system a useful data set validation tool as well. In particular, we found a significant amount of errors in the original CLUTRR dataset that are hard to detect manually. 


While the new version of GPT-3 \cite{brown20language} (text-davinci-003) shows improvement over its predecessors, we observe that it also retains critical limitations.  In the process, we develop prompt methods for semantic parsing to overcome some of them.

\BOCC
In summary, our contributions are as follows. 
\begin{itemize}
\item We present a dual-process model for robust and general question-answering by combining few-shot semantic parsing of LLMs and symbolic logic programming, which complement each other's weaknesses. Our method does not require training on the dataset. 

\item We propose prompt methods for semantic parsing and evaluate them on multiple benchmarks. We identify a few critical limitations with using GPT-3 for this purpose.

\item  We show that our method achieves state-of-the-art performance on several benchmarks. 

\end{itemize}
\EOCC

The implementation of our method is publicly available
online at 
\url{https://github.com/azreasoners/LLM-ASP}. 

\section{Preliminaries}  \label{sec:prelim}

\subsection{Semantic Parsing and LLMs}

Semantic parsing involves converting a natural language query or statement into a structured representation that a computer can understand and manipulate. 
Statistical methods have increased in popularity \cite{zelle96learning, miller96fully,zettlemoyer05learning,wong07learning}, and encoder-decoder models in particular have been widely used \cite{dong16language, jia16data, kovcisky16semantic}. However, these statistical methods require annotated input and output pairs. Furthermore, machine learning models 
often fail to compositionally generalize to unseen data \cite{lake18generalization}. 

{More recently, pre-trained language models have been applied to semantic parsing tasks 
 \cite{liu21pre}, such as generating SQL queries, SPARQL queries, logical forms, or programs, from natural language,
 together with fine-tuning or prompt-tuning on pre-trained models, such as BART, RoBERTa and GPT-2 \cite{chen20low,shin21constrained,schucher22power}.
With larger pre-trained networks, such as GPT-3, prompting appears to yield a reasonable semantic parser without the need for fine-tuning \cite{shin21constrained, drozdov22compositional}.

Another line of related work is to apply pre-trained language models to relation extraction, the task of extracting semantic relationships from a text given two or more entities \cite{liu21pre}.
\citet{wang22ielm} do zero-shot relation extraction with pre-trained language models from the BERT family and GPT-2 variants.
\citet{zhou-chen-22-improved} fine-tune BERT and RoBERTa models for the extraction of sentence-level relations. \citet{chen22knowprompt} apply prompt-tuning to RoBERT\_LARGE for relation extraction. 
Similar to ours, \citet{agrawal22large} use a few-shot prompt with GPT-3 for the extraction of clinical relations.

\subsection{Dual-System Model} 
                                               
There is increasing interest in combining neural and symbolic systems \citep{marcus18deep,lamb20graph,sarker21neuro}. 
Such dual-system models achieved new state-of-the-art results in visual question answering 
\cite{goldman18weakly,sampat18model,yi19clevrer,chen20grounding,ding21dynamic}.
In the case of textual problems, to improve LLMs to generate more consistent and coherent sentences, \citet{nye21improving} suggest that generation be decomposed into two parts: candidate sentence generation by an LLM (system 1 thinking) and a logical pruning process (system 2 thinking) implemented via a separate symbolic module. They demonstrate that this neuro-symbolic, dual-process model requires less training data, achieves higher accuracy, and exhibits better generalization. 
However, the main limitation of their work is that the symbolic module is manually constructed in Python code for the specific task at hand, requiring subtantial efforts. Additionally, their Python symbolic module is not readily reusable or composable. Furthermore, their main results primarily focus on the problem of consistent text generation, rather than evaluating the method on the datasets and comparing it with existing models. This is because writing the world models in Python is not a scalable approach.  
    
In our work, we follow the idea presented in~\cite{nye21improving} but adopt logic programming in place of the System 2 process. We argue that this combination is much more appealing than the approach in~\cite{nye21improving}, as it can achieve the promised results without the limitations mentioned above.

\subsection{Answer Set Programming}

Answer Set Programming (ASP) \cite{lif08,bre11} is a declarative logic programming paradigm that has been shown to be effective in knowledge-intensive applications. It is based on the stable model (a.k.a. answer set) semantics of logic programs \cite{gel88}, which could express causal reasoning, default reasoning, aggregates, and various other constraints. There are several efficient solvers, such as {\sc clingo}, {\sc dlv}, and {\sc wasp}.  
We use {\sc clingo} v5.6.0 as the answer set solver. 
For the language of {\sc clingo}, we refer the reader to the textbook \cite{lifschitz19answer} or the {\sc clingo} manual.\footnote{\url{https://github.com/potassco/guide/releases}.}

It is also known that classical logic-based action formalisms, such as the situation calculus \cite{mcc69,rei01} and the event calculus \cite{shan95}, can be formulated as answer set programs. 
For example, the following is one of the axioms in Discrete Event Calculus stating the commonsense law of inertia, saying that fluent $F$ holds at the next time if there is no action affecting it.
\begin{lstlisting}
% (DEC5)
holds_at(F,T+1) :- timepoint(T), fluent(F),
    holds_at(F,T), -released_at(F,T+1), 
    not terminated(F,T).
\end{lstlisting}
Such a rule is universal and applies to almost all objects.

Answer set programs are also known to be {\em elaboration tolerant} \cite{mccarthy98elaboration}.
There has been work on modularizing knowledge bases in ASP, such as module theorem \cite{oik06,babb12module} and knowledge modules \cite{baral06macros}. 
While ASP has been widely applied to many reasoning problems, it has not been considered as much in reasoning with natural language text because its input is expected to be strictly in a logical form, giving little flexibility in accepting diverse forms of natural language input. 



\section{Our Method} \label{sec:method}

\begin{figure*}[ht]
\begin{center}
\includegraphics[width=1.6\columnwidth]{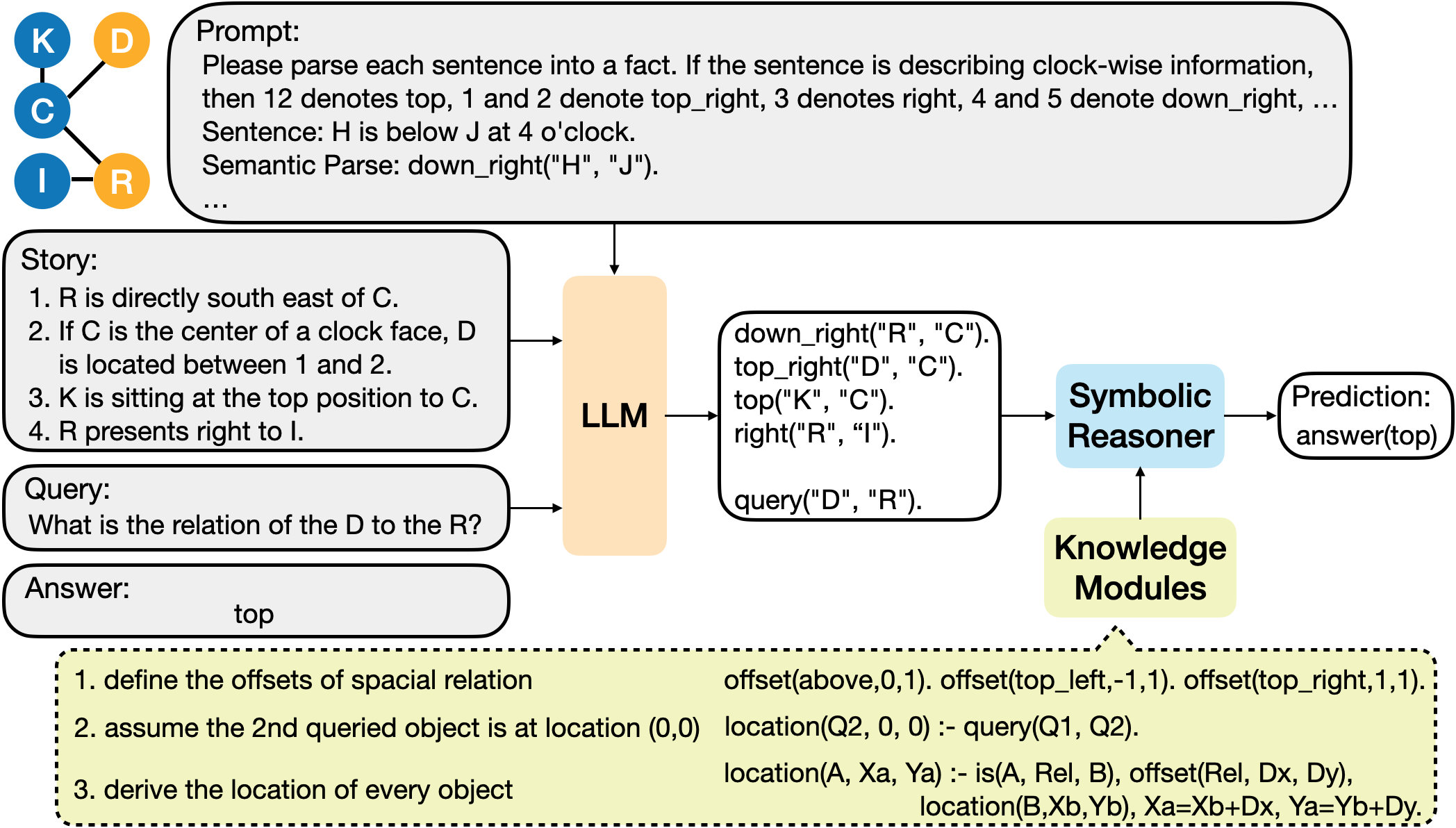}
\end{center}
\caption{The GPT-3+ASP pipeline for the StepGame dataset.}
\label{fig:pipeline}
\end{figure*}

We refer to our framework as [LLM]+ASP where [LLM] denotes a large pre-trained network such as GPT-3, which we use as a semantic parser to generate input to the ASP reasoner. 
Specifically, we assume data instances of the form $\langle S, q, a \rangle$, where $S$ is a context story in natural language, $q$ is a natural language query associated with $S$, and $a$ is the answer.
We use an LLM to convert a problem description (that is, context $S$ and query $q$) into atomic facts, which are then fed into the ASP solver along with background knowledge encoded as ASP rules. 
The output of the ASP solver is interpreted as the prediction for the given data instance.
Figure~\ref{fig:pipeline} illustrates the inference flow in the context of StepGame. 
The pipeline is simple but general enough to bke applied to various tasks without the need for retraining. It only requires replacing the few-shot prompts to the LLM and the ASP background knowledge with those suitable for the new tasks. 

By combining LLMs and ASP in this manner, we enable robust symbolic reasoning that can handle diverse and unprocessed textual input. 
The ASP knowledge modules remain unaffected by the diverse forms of input text that express the same facts. Our method does not rely on training datasets. Instead, a few examples that turn natural language sentences into atomic facts are sufficient to build a semantic parser due to the learned representations in LLMs. Furthermore, ASP knowledge modules can be reused for different tasks.

\subsection{Prompts for Fact Extraction} \label{subsec:prompt_for_symbol_grounding}



We use GPT-3 to extract atomic facts from the story and query. Most of the time, giving several examples yields accurate semantic parsing. 
The following is an example prompt for bAbI. 
\begin{lstlisting}
Please parse the following statements into facts. The available keywords are: pickup, drop, and go.
Sentence: Max journeyed to the bathroom.
Semantic parse: go(Max, bathroom).

Sentence: Mary grabbed the football there.
Semantic parse: pickup(Mary, football).
...
\end{lstlisting}

We find that GPT-3 is highly tolerable to linguistic variability.
For example, in StepGame, GPT-3 can turn various sentences below into the same atomic fact {\tt top\_right("C","D")}.
\begin{lstlisting}
C is to the top right of D.
C is to the right and above D at an angle of about 45 degrees.
C is at a 45 degree angle to D, in the upper righthand corner.
C is directly north east of D.
C is above D at 2 o'clock.
\end{lstlisting}

In the experiments to follow, we find that the following strategy works well for fact extraction.
\begin{enumerate}
\item 
In general, we find that 
if the information in a story (or query) can be extracted independently, parsing each sentence separately (using the same prompt multiple times) typically works better than parsing the whole story. 

\item  
There is certain commonsense knowledge that GPT-3 is not able to leverage from the examples in the prompt. In this case, detailing the missing knowledge in the prompt could work.
For example, in StepGame, clock numbers are used to denote cardinal directions, but GPT-3 couldn't translate correctly even with a few examples in the prompt. It works after enumerating all cases (``12 denotes top, 1 and 2 denote top\_right, 3 denotes right, $\dots$'') in the prompt.
%


\item
Semantic parsing tends to work better if we instruct GPT-3 to use a predicate name that better reflects the intended meaning of the sentence. For example, "A is there and B is at the 5 position of a clock face" is better to be turned into {\tt down\_right(B,A)} than {\tt top\_left(A,B)} although, logically speaking, the relations are symmetric. 

\end{enumerate}
The complete set of prompts for semantic parsing is given in Appendix~\ref{app:sec:prompts}.

\subsection{Knowledge Modules}

Instead of constructing a minimal world model for each task in Python code \cite{nye21improving}, we use ASP knowledge modules.
While some knowledge could be lengthy to be described in English, it could be concisely expressed in ASP. 
For example, the {\bf location} module contains rules for spatial reasoning in a 2D grid space and is used for bAbI, StepGame, and gSCAN. 
Below is the main rule in the {\bf location} module that computes the location {\tt (Xa,Ya)} of object {\tt A} from the location {\tt (Xb,Yb)} of object {\tt B} by adding the offsets {\tt (Dx,Dy)} defined by the spatial relation {\tt R} between {\tt A} and {\tt B}.
\begin{lstlisting}
location(A, Xa, Ya) :- location(B, Xb, Yb), 
    is(A, R, B), offset(R, Dx, Dy),
    Xa=Xb+Dx, Ya=Yb+Dy.
\end{lstlisting}
The {\bf location} module also includes 9 predefined offsets, e.g., {\tt offset(left,-1,0)}, that can be used to model multi-hop spatial relations of objects or effects of a robot's moving in a 2D space.
For example, queries in StepGame are about the spatial relation {\tt R} of object {\tt A} to {\tt B}. Using the {\bf location} module, one can fix {\tt B}'s location to be {\tt (0,0)} and compute the spatial relation {\tt R} based on the location of {\tt A} as follows.
%
\begin{lstlisting}
location(B, 0, 0) :- query(A, B).
answer(R) :- query(A, B), location(A, X, Y),
    offset(R, Dx, Dy),
    Dx=-1: X<0; Dx=0: X=0; Dx=1: X>0;
    Dy=-1: Y<0; Dy=0: Y=0; Dy=1: Y>0.
\end{lstlisting}
The second rule above contains six {\em conditional literals} among which {\tt Dx=-1:X<0} says that ``{\tt Dx} must be -1 if {\tt X<0}.'' For example, 
if {\tt A}'s location {\tt (X,Y)}  is {\tt (-3,0)}, then {\tt (Dx,Dy)} is {\tt (-1,0)} and the answer {\tt R} is {\tt left}.
Similar rules can also be applied to bAbI task 17, which asks if {\tt A} is {\tt R} of {\tt B}.


In the above rules, the relation {\tt R} in, e.g., {\tt is(A,R,B)}, is a variable and can be substituted by any binary relation. 
Such high-order representation turns out to be quite general and applicable to many tasks that query relation or its arguments. 
%


\begin{figure}[h]
\begin{center}
\includegraphics[width=0.9\columnwidth]{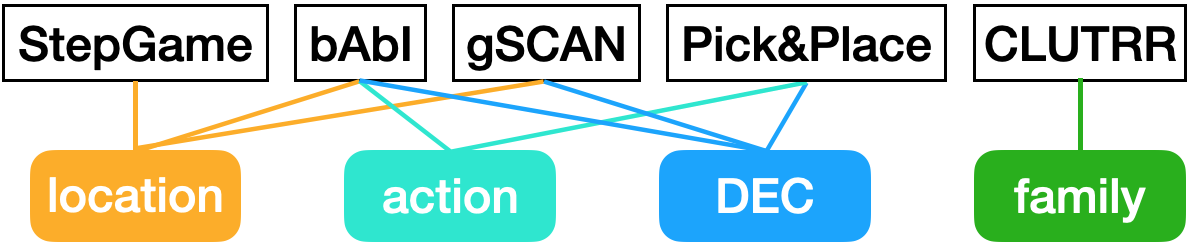}
\end{center}
\caption{The knowledge modules at the bottom are used in each task on the top.}
\label{fig:domain2modules}
\end{figure}

Figure~\ref{fig:domain2modules} shows the knowledge modules used in this paper, where {\bf DEC} denotes the Discrete Event Calculus axioms from \cite{mue06,lee12reformulating}. 
In this section, we explained the main rules in the {\bf location} module. The complete ASP knowledge modules are given in 
 Appendix~\ref{app:sec:asp-modules}.


\begin{table*}[!ht]

\begin{center}
\footnotesize
\begin{tabular}{| l | c | c | c | c | c | c |}
\hline
\textbf{Task}  

                           & {\textbf{GPT-3(d3)}} & {\textbf{GPT-3(d3)}} &  {\textbf{GPT-3(d3)}} & {\textbf{STM}\cite{le20self}} & \multicolumn{2}{c|}{\textbf{QRN}\cite{seo16query}}   \\
                           & Few-Shot                        & CoT   & +ASP &(10k train) & (10k train) & (1k train)
\\
\cline{1-7}
1: Single supporting fact  & 98.4    &   97.3     & 100.0 & 100.0 $\pm$ 0.0    &  100.0 & 100.0 \\ \hline
2: Two supporting facts    & 60.8    &   72.2     & 100.0 & 99.79 $\pm$ 0.23   &  100.0 & 99.3  \\ \hline
3: Three supporting facts  & 39.6    &   54.1     & 100.0 & 97.87 $\pm$ 1.14   &  100.0 & 94.3  \\ \hline
4: Two arg relations       & 60.4    &   72.7     & 100.0 & 100.0 $\pm$ 0.0    &  100.0 & 100.0 \\ \hline
5: Three arg relations     & 88.2    &   89.1     & 99.8  & 99.43 $\pm$ 0.18   &  100.0 & 98.9  \\ \hline
6: Yes/no questions        & 97.4    &   97.3     & 100.0 & 100.0 $\pm$ 0.0    &  100.0 & 99.1  \\ \hline
7: Counting                & 90.6    &   88.6     & 100.0 & 99.19 $\pm$ 0.27   &  100.0 & 90.4  \\ \hline
8: Lists/sets              & 96.2    &   97.1     & 100.0 & 99.88 $\pm$ 0.07   &  99.6  & 94.4  \\ \hline
9 : Simple negation        & 98.4    &   98.2     & 100.0 & 100.0 $\pm$ 0.0    &  100.0 & 100.0 \\ \hline
10: Indefinite knowledge   & 93.6    &   92.4     & 100.0 & 99.97 $\pm$ 0.06   &  100.0 & 100.0 \\ \hline
11: Basic coreference      & 93.6    &   99.2     & 100.0 & 99.99 $\pm$ 0.03   &  100.0 & 100.0 \\ \hline
12: Conjunction            & 88.6    &   88.8     & 100.0 & 99.96 $\pm$ 0.05   &  100.0 & 100.0 \\ \hline
13: Compound coreference   & 98.4    &   97.3     & 100.0 & 99.99 $\pm$ 0.03   &  100.0 & 100.0 \\ \hline
14: Time reasoning         & 78.0    &   91.5     & 100.0 & 99.84 $\pm$ 0.17   &  99.9  & 99.2  \\ \hline
15: Basic deduction        & 57.0    &   95.0     & 100.0 & 100.0 $\pm$ 0.0    &  100.0 & 100.0 \\ \hline
16: Basic induction        & 90.8    &   97.5     & 100.0 & 99.71 $\pm$ 0.15   &  100.0 & 47.0  \\ \hline
17: Positional reasoning   & 66.0    &   70.8     & 100.0 & 98.82 $\pm$ 1.07   &  95.9  & 65.6  \\ \hline
18: Size reasoning         & 89.8    &   97.1     & 100.0 & 99.73 $\pm$ 0.28   &  99.3  & 92.1  \\ \hline
19: Path finding           & 21.0    &   28.7     & 100.0 & 97.94 $\pm$ 2.79   &  99.9  & 21.3  \\ \hline
20: Agents motivations     & 100.0   &   100.0    & 100.0 & 100.0 $\pm$ 0.0    &  100.0 & 99.8  \\ \hline \hline
Average                    & 80.34   &   86.18    & {\bf 99.99} & 99.85              & 99.70  & 90.1  \\  
\bottomrule

\end{tabular}
\end{center}
\caption{Test accuracy on 20 tasks in bAbI data} 
\label{tb:babi_all}
\end{table*}

\section{Experiments} \label{sec:experiments}

We apply the method in the previous section to four datasets.\footnote{Due to space restriction, we put the experiments about Pick\&Place in Appendix~\ref{app:sec:robot}.} 
As a reminder, our approach involves few-shot in-context learning and does not require training. We use the same pipeline as shown in Figure~\ref{fig:pipeline}, but with different prompts and knowledge modules for each dataset. 
For more detailed information about the experimental settings, please refer to the appendix.


\subsection{bAbI} \label{ssec:babi-experiment}

The bAbI dataset \cite{weston15towards} is a collection of 20 QA tasks that have been widely applied to test various natural language reasoning problems, such as deduction, path-finding, spatial reasoning, and counting. 
State-of-the-art models, such as self-attentive associative-based two-memory model (STM) \cite{le20self} and Query-Reduction networks (QRN) \cite{seo16query} achieve close to 100\% accuracy after training with 10k instances while QRN's accuracy drops to 90\% with 1k training instances.

We first designed two GPT-3 baselines, one with few shot prompts (containing a few example questions and answers)  and the other with Chain-of-Thought (CoT) prompts \cite{wei22chain}, which state the relevant information to derive the answer.

We also apply GPT-3+ASP. 
For example, we use GPT-3 to turn ``the kitchen is south of the bathroom'' into an atomic fact {\tt is(kitchen, southOf, bathroom)} by giving a few examples of the same kind.
Regarding knowledge modules, Tasks 1--3, 6--9, 10--14, and 19 are about events over time and use the \textbf{DEC} knowledge module. Tasks 4, 17, and 19 require various domain knowledge modules such as \textbf{location} and \textbf{action} knowledge modules. The remaining tasks do not require domain knowledge and rely only on simple rules to extract answers from parsed facts. 

Table~\ref{tb:babi_all} compares our method with the two GPT-3 baselines, as well as two state-of-the-art methods on bAbI datasets, STM and QRN. 
Interestingly, the new GPT-3, text-davinci-003 (denoted GPT-3 (d3)), with basic few-shot prompting achieves 80.34\% accuracy, while CoT improves it to 86.18\%. GPT-3(d3)+ASP achieves state-of-the-art performance on bAbI with 99.99\% average performance among all tasks, producing only two answers that disagree with the labels in the dataset. It turns out that the two questions are malformed since the answers are ambiguous, and our model's answers can be considered correct.\footnote{See Appendix~\ref{appendix:Dataset_errors:bAbI} for the examples.} 

\subsection{StepGame} \label{ssec:stepgame-experiment}

Although bAbI has been extensively tested, it has several problems. 
\citet{shi22stepgame} note data leakage between the train and the test sets where named entities are fixed and only a small number of relations are used.
\citet{palm18recurrent} point out that models do not need multi-hop reasoning to solve the bAbI dataset.
To address the issues, \citet{shi22stepgame} propose the StepGame dataset.  It is a contextual QA dataset in which the system is required to interpret a story $S$ about spatial relationships among several entities and answers a query $q$ about the relative position of two of those entities, as illustrated in Figure~\ref{fig:pipeline}. 
Unlike the bAbI dataset, StepGame uses a large number of named entities, and requires multi-hop reasoning up to as many as 10 reasoning steps. 

In the basic form of the StepGame dataset, each story consists of $k$ sentences that describe $k$ spatial relationships between $k + 1$ entities in a chain-like shape.
In this paper, we evaluate the StepGame dataset with noise, where the original chain is extended with noise statements by branching out with new entities and relations. 
 
Similarly to bAbI, we designed two GPT-3 baselines and applied our method to the StepGame data set. More details on the prompts are available in Appendix~\ref{appendix:prompt:stepgame}.

\begin{table}[ht!]
	\centering
	{\small 
	\begin{tabular}{l|ccccc}
		\toprule
		 Method & k=1 & k=2 & k=3 & k=4 & k=5 \\
		\midrule
		RN & 22.6 & 17.1 & 15.1 & 12.8 & 11.5 \\
		RRN & 24.1 & 20.0 & 16.0 & 13.2 & 12.3 \\
		UT & 45.1 & 28.4 & 17.4 & 14.1 & 13.5 \\
		STM & 53.4 & 36.0 & 23.0 & 18.5 & 15.1 \\
		TPR-RNN & 70.3 & 46.0 & 36.1 & 26.8 & 24.8 \\
		TP-MANN & 85.8 & 60.3 & 50.2 & 37.5 & 31.3 \\
		SynSup & {\bf 98.6} & {\bf 95.0} & {\bf 92.0} & 79.1 & 70.3 \\
		\midrule
        Few-Shot (d3) & 55.0 & 37.0  & 25.0 & 30.0 & 32.0 \\
        CoT (d3) & 61.0  & 45.0  & 30.0 & 35.0 & 35.0 \\
        GPT-3(c1)+ASP & 44.7 & 38.8 & 40.5 & 58.8 & 62.4 \\
		GPT-3(d2)+ASP & 92.6 & 89.9 & 89.1 & {\bf 93.8} & {\bf 92.9} \\
 		\bottomrule
		\toprule
		Method & k=6 & k=7 & k=8 & k=9 & k=10 \\
		\midrule
		RN & 11.1 & 11.5 & 11.2 & 11.1 & 11.3 \\
		RRN & 11.6 & 11.4 & 11.8 & 11.2 & 11.7 \\
		UT & 12.7 & 12.1 & 11.4 & 11.4 & 11.7 \\
		STM & 13.8 & 12.6 & 11.5 & 11.3 & 11.8 \\
		TPR-RNN & 22.3 & 19.9 & 15.5 & 13.0 & 12.7 \\
		TP-MANN & 28.5 & 26.5 & 23.7 & 22.5 & 21.5 \\
		SynSup & 63.4 & 58.7 & 52.1 & 48.4 & 45.7 \\
		\midrule
        Few-Shot (d3) & 29.0 & 21.0 & 22.0 & 34.0 & 31.0 \\
        CoT (d3) & 27.0 & 22.0 & 24.0 & 23.0 & 25.0 \\
        GPT-3(c1)+ASP & 57.4 & 56.2 & 58.0 & 56.5 & 54.1 \\
		GPT-3(d2)+ASP & {\bf 91.6} & {\bf 91.2} & {\bf 90.4} & {\bf 89.0} & {\bf 88.3} \\
		\bottomrule
	\end{tabular}
	}
	\caption{Test accuracy on the StepGame test dataset, where 
 (c1), (d2), and (d3) denote text-curie-001, text-davinci-002, and text-davinci-003 models, respectively}
	\label{tb:stepgame}
\end{table}

For each $k\in \{1,\dots,10\}$, the StepGame dataset with noise consists of 30,000 training samples, 1000 validation samples, and 10,000 test samples.
To save the API cost for GPT-3, we only evaluated the two GPT-3 baselines on the first 100 test samples and evaluated our method on the first 1,000 test samples for each $k\in \{1,\dots,10\}$. 
Table~\ref{tb:stepgame} compares the accuracy of our method with the two baselines of GPT-3 and the current methods, i.e. RN \citep{santoro17asimple}, RRN \citep{palm18recurrent}, UT \citep{dehghani18universal}, STM \citep{le20self}, TPR-RNN \citep{schlag18learning}, TP-MANN \citep{shi22stepgame}, and SynSup (with pre-training on the SPARTUN dataset) \cite{mirzaee22transfer}.
Surprisingly, the GPT-3 baselines could achieve accuracy comparable to other models (except for SynSup) for large $k$ values. CoT does not always help and decreases the accuracy with big $k$s. This may be because there is a higher chance of making a mistake in a long chain of thought.
GPT-3(d2)+ASP outperforms all state-of-the-art methods and the GPT-3 baselines by a large margin for $k=4,\dots,10$. 
Although SynSup achieves a higher accuracy for $k=1,2,3$, this is misleading due to errors in the dataset. As we analyze below, about 10.7\% labels in the data are wrong. The SynSup training makes the model learn to make the same mistakes over the test dataset, which is why its performance looks better than ours. 

}


The modular design of GPT-3+ASP enables us to analyze the reasons behind its wrong predictions. We collected the first 100 data instances for each $k\in\{1,\dots,10\}$ and manually analyzed the predictions on them. 
Among 1000 predictions of GPT-3(d2)+ASP, 108 of them disagree with the dataset labels, and we found that 107 of those have errors in the labels. For example, given the story and question ``{\em J and Y are horizontal and J is to the right of Y. What is the relation of the agent Y with the agent J?}'', the label in the dataset is ``right'' while the correct relation should be ``left''.%
\footnote{The remaining disagreeing case is due to text-davinci-002's mistake.  For the sentence, ``{\em if E is the center of a clock face, H is located between 2 and 3}.'' text-davinci-002 turns it into  ``right(H, E)'' whereas text-davinci-003 turns it into ``top-right(H, E)'' correctly. To save API cost for GPT-3, we did not re-run the whole experiments with text-davinci-003.}
Recall that our method is interpretable, so we could easily identify the source of errors. 


\subsection{CLUTRR} \label{ssec:CLUTRR-experiment}

CLUTRR \citep{sinha19clutrr} is a contextual QA dataset that requires inferring family relationships from a story. Sentences in CLUTRR are generated using 6k template narratives written by Amazon Mechanical Turk crowd-workers, and thus are more realistic and complex compared to those in bAbI and StepGame.
\NBB{Mention mechanical turk here and why this is more realistic. Done.}

CLUTRR consists of two subtasks, {\em systematic generalization} that evaluates stories containing unseen combinations of logical rules \citep{minervini20learning,bergen21systematic} and {\em robust reasoning} that evaluates stories with noisy descriptions \citep{tian21generative}.
Since we use ASP for logical reasoning, which easily works for any combination of logical rules, we focus on the robust reasoning task. 




\begin{table}[ht!]
	\centering
	{\small 
	\begin{tabular}{l|l|cccc}
		\toprule
		 Method & CLU. & clean & supp. & irre. & disc. \\
		\midrule
        RN & 1.0 & 49 & 68 & 50 & 45 \\
        MAC &  1.0 & 63 & 65 & 56 & 40 \\
        Bi-att & 1.0 & 58 & 67 & 51 & 57 \\
		GSM &  1.0 & {\bf 68.5} & 48.6 & 62.9 & 52.8 \\
        \hline
        GPT-3(d3)+ASP &  1.0 & {\bf 68.5} & {\bf 82.8} & {\bf 74.8} & {\bf 67.4} \\
        \midrule
	GPT-3(d3)+ASP &  1.3 & 97.0 & 84.0 & 92.0 & 90.0 \\
		\bottomrule
	\end{tabular}
	}
 \caption{Test accuracy on 4 categories in CLUTRR 1.0 and CLUTRR 1.3 datasets} 
  \vspace{-1em}
	\label{tb:clutrr_1.3}
\end{table}

Table~\ref{tb:clutrr_1.3} compares our method with RN \citep{santoro17asimple}, MAC \cite{hudson18compositional}, BiLSTM-attention \cite{sinha19clutrr}, and GSM \citep{tian21generative} on the original CLUTRR dataset, namely CLUTRR 1.0, in four categories of data instances: clean, supporting, irrelevant, and disconnected \citep{sinha19clutrr}. 
Except for our method, all other models are trained on the corresponding category of CLUTRR training data.
Although our method achieves similar or higher accuracies in all categories, they are still much lower than we expected. 

We found that such low accuracy is due to the clear errors in CLUTRR, originating mostly from errors in the template narratives or the generated family graphs that violate common sense. The authors of CLUTRR recently published CLUTRR 1.3 codes to partially resolve this issue.~\footnote{\url{https://github.com/facebookresearch/clutrr/tree/develop}}
With the new code, we created a new dataset, namely CLUTRR 1.3, consisting of $400$ data instances with $100$ for each of the four categories. The last row in Table~\ref{tb:clutrr_1.3} shows that our method actually performs well on realistic sentences in CLUTRR. 
Indeed,
with our method (using text-davinci-003) on CLUTRR 1.3 dataset, 363 out of 400 predictions are correct, 16 are still wrong due to data mistakes (e.g., the label says ``Maryann has an uncle Bruno'' while the noise sentence added to the story is ``Maryann told her son Bruno to give the dog a bath''), and 21 are wrong due to GPT-3's parsing mistakes (e.g., GPT-3 turned the sentence ``Watt and Celestine asked their mother, if they could go play in the pool'' into {\tt mother("Watt", "Celestine")}. Since the sentences in CLUTRR 1.3 are more realistic than those in bAbI and StepGame, GPT-3 makes more mistakes even after reasonable efforts of prompt engineering.
More details on data errors and GPT-3 errors are available in Appendix~\ref{appendix:data_errors:clutrr} and Appendix~\ref{app:sec:gpt3-errors}.

\begin{table}[ht!]
	\centering
	{\small 
	\begin{tabular}{l|cccc}
		\toprule
		 Method &  clean & supp. & irre. & disc. \\
		\midrule
        DeepProbLog &  100 & 100 & 100 & 94 \\
        \hline
        GPT-3(d2)+ASP &  100 & 100 & 97 & 97 \\
        GPT-3(d3)+ASP &  100 & 100 & 100 & 100 \\
		\bottomrule
	\end{tabular}
	}
 \caption{Test accuracy on CLUTRR-S dataset} 
  \vspace{-1em}
  \label{tb:clutrr-s}
\end{table}

We also evaluated our method on a simpler and cleaner variant of the CLUTRR data set, namely CLUTRR-S, that was used as a benchmark problem for a state-of-the-art neuro-symbolic approach DeepProbLog \citep{manhaeve21neural}.
Table~\ref{tb:clutrr-s} compares the accuracy of our method and DeepProbLog in all 4 categories of test data. GPT-3(d3)+ASP achieves 100\% accuracy, outperforming DeepProbLog without the need for training.

\medskip
\noindent{\bf Remark:~} Due to the modular structure, our method could serve as a data set validation tool to detect errors in a dataset. We detected 107 wrong data instances in the first 1000 data in StepGame and 16 wrong data instances in the 400 data in CLUTRR 1.3.




\subsection{gSCAN} \label{ssec:gscan}

The gSCAN dataset \cite{ruis20benchmark} poses a task in which an agent must execute action sequences to achieve a goal (specified by a command in a natural language sentence) in a grid-based visual navigation environment. The dataset consists of two tasks, and we evaluate our method on the data splits from the compositional generalization task. There is one shared training set, one test set (split A) randomly sampled from the same distribution of the training set, and seven test sets (splits B to H) with only held-out data instances (i.e., not appearing in the training set) in different ways. 

In the gSCAN dataset, each data instance is a tuple $\langle G, q, a \rangle$ where $G$ is the grid configuration (in JSON format) describing the size of the gird, the location and direction of the agent, and the location and features of each object in the grid; $q$ is a query (e.g., ``pull a yellow small cylinder hesitantly''); and $a$ is the answer in the form of a sequence of actions (e.g., ``turn right, walk, stay, pull, stay, pull, stay''). 
For each data instance, we (i) use a Python script to extract atomic facts (e.g., {\tt pos(agent,(2,3))}) from the grid configuration $G$; (ii) extract atomic facts from query $q$ into atomic facts (e.g., {\tt query(pull)}, {\tt queryDesc(yellow)}, {\tt while(hesitantly)}) using GPT-3; and (iii) predict the sequence of actions for this query using ASP. The details of the prompts are given in Appendix~\ref{appendix:prompt:gscan}.

\begin{table}[ht!]
	\centering
	{\small 
	\begin{tabular}{l|cccc}
		\toprule
		 Method & A & B & C & D \\
		\midrule
		GECA & 87.60 & 34.92 & 78.77 & 0.00 \\
		DualSys & 74.7 & 81.3 & 78.1 & 0.01 \\
		Vilbert+CMA & 99.95 & 99.90 & 99.25 & 0.00 \\
		\midrule
        GPT-3(c1)+ASP & 98.30 & {\bf 100} & {\bf 100} & {\bf 100} \\
        GPT-3(d2)+ASP & {\bf 100} & {\bf 100} & {\bf 100} & {\bf 100} \\
 		\bottomrule
		\toprule
		Method & E & F & G & H \\
		\midrule
		GECA & 33.19 & 85.99 & 0.00 & 11.83 \\
		DualSys & 53.6 & 76.2 & 0.0 & 21.8 \\
		Vilbert+CMA & 99.02 & 99.98 & 0.00 & 22.16 \\
		\midrule
        GPT-3(c1)+ASP & {\bf 100} & {\bf 100} & {\bf 100} & {\bf 100} \\
        GPT-3(d2)+ASP & {\bf 100} & {\bf 100} & {\bf 100} & {\bf 100} \\
		\bottomrule
	\end{tabular}
	}
	\caption{Test accuracy on the gSCAN dataset} 
 \vspace{-1em}
	\label{tb:gscan}
\end{table}

Table~\ref{tb:gscan} compares the accuracy of our method and the state-of-the-art methods, i.e., GECA \cite{ruis20benchmark}, DualSys \citep{nye21improving} and Vilbert+CMA \cite{qiu21systematic}, on the gSCAN test dataset in eight splits. 
To save API cost for GPT-3,
we only evaluated the first 1000 data instances of each split. With text-davinci-002, our method GPT-3+ASP achieves 100\% accuracy. 
With text-curie-001, the accuracy is slightly lower, making 17 errors in split A. The errors are of two kinds. The language model fails to extract adverbs in the correct format for 11 data instances (e.g., GPT-3 responded {\tt queryDesc(while spinning)} instead of {\tt while(spinning)}) and didn't ground the last word in a query for 6 data instances (e.g., for query {\tt walk to a small square}, GPT-3 missed an atomic fact {\tt queryDesc(square)}). 
Once the parsed results are correct, ASP does not make a mistake in producing plans.

\subsection{Findings}
The following summarizes the findings of the experimental evaluation. 

\begin{itemize}
\item  Our experiments confirm that LLMs like GPT-3 are still not good at multi-step reasoning despite various prompts we tried. Chain-of-Thought is less likely to improve accuracy when a long chain of thought is required. 

\item On the other hand, LLMs are surprisingly good at turning a variety of expressions into a "canonical form" of information extraction. This in turn allows ASP knowledge modules to be isolated from linguistic variability in the input. 

\item  Even for generating simple atomic facts, larger models tend to perform better. For example, in StepGame and gSCAN, text-curie-001 performs significantly worse compared to text-davinci-002 (Tables~\ref{tb:stepgame} and \ref{tb:gscan}).

\item  The total amount of knowledge that needs to be encoded for all of the above datasets is not too large. This is in part due to the fact that GPT-3 "normalized" various forms of input sentences for ASP to process and that knowledge modules could be reused across different datasets. 

\item The modular design of our approach makes it possible to locate the root cause of each failed prediction in the training data and improve upon it. There are three sources of errors: semantic parsing in LLMs, symbolic constraints, and the dataset itself, and we can resolve the first two issues by improving the prompts and updating the constraints, respectively.

\item Our framework could serve as a few-shot dataset justifier and corrector. Among all predictions by our method that do not align with the labels, almost all of them (with only a few exceptions discussed in the paper) are due to errors in the dataset.

\end{itemize}

\BOC
{\bf Findings}
\begin{itemize}
\item  We see curie models do not perform well. [ eveidence; which dataset clearly distinguishes between curie vs. davinci? Do we need a large LM for semantic parsing or simpler is okay  ]
\item  Unlike previous works, LLMs are very adaptive to the various neuance, thus the main ASP program does not need change. 
\end{itemize}
\EOC

\section{Conclusion} \label{sec:conclusion}

Symbolic logic programming was previously considered limited in its ability to reason from text due to its inability to handle various and ambiguous linguistic expressions. However, combining it with a large language model that has learned distributed representations helps alleviate this problem. The method not only achieves high accuracy but also produces interpretable results, as the source of the errors can be identified. 
It is also general; by using pre-trained networks with few-shot prompts and reusable knowledge modules, adapting to a new domain does not require extensive training. 
 
The knowledge modules used in our experiments are reusable. 
For the above experiments, the modules are relatively simple to write, as are the prompts for parsing natural language for LLMs. However, acquiring this kind of knowledge on a massive scale is also an important line of research \cite{liu04conceptnet, bosselut19comet,hwang21comet} that needs to be combined. 
In addition, it is possible to use LLM's code generation capability \cite{chen21evaluating} to generate logic program rules, which we leave for future work.

One may think that the logic rules are too rigid. However, there are many weighted or probabilistic rules that can be defeated \cite{richardson06markov,fierens13inference,lee18weight}. They could be used for more realistic settings, but for the benchmark problems above, they were not needed.

\section*{Ethical Considerations}
All datasets used in this paper are publicly available.
For CLUTRR dataset, the gender information is essential to tell if, e.g., A is B's uncle or niece. We used GPT-3 to predict the genders of persons in each story. Since each story is systematically generated using sampled common first names and sampled sentence templates, it does not reveal any identity.
As mentioned, the original CLUTRR dataset had
some errors, and we describe carefully the codes and settings of the generated CLUTRR 1.3 dataset in Appendix~\ref{app:subsec:clutrr_gen}.

\section*{Limitations}
The current work requires that knowledge modules be written by hand. Commonly used axioms, such as general knowledge like the commonsense law of inertia expressed by event calculus, can be reused easily, but there are vast amounts of other commonsense knowledge that are not easy to obtain. LLMs could be used to supply this information, but we have not tried.
Knowledge graphs, such as ConceptNet \cite{liu04conceptnet}, COMET \cite{bosselut19comet} and ATOMIC \cite{hwang21comet}, can be utilized to populate ASP rules. 
Like code models, we expect that LLMs could generate ASP code, which we leave for future work. 

Also, when using large language models, despite various efforts, sometimes it is not understandable why they do not behave as expected. 

\section*{Acknowledgements} 

This work was partially supported by the National Science Foundation under Grant IIS-2006747.


\bibliographystyle{acl_natbib}

\newpage

\appendix


{\bf \Large Appendix}
\medskip



Section~\ref{app:sec:robot} presents another experiment with robot planning. 
Section~\ref{app:sec:clutrr} discusses more details about how we generated CLUTRR dataset and the experimental result on CLUTRR 1.0. 
Section~\ref{app:sec:prompts} presents GPT-3 prompts for semantic parsing.
Section~\ref{app:sec:gpt3-errors} enumerates the errors with GPT-3 in semantic parsing. 
Section~\ref{app:sec:asp-modules} presents ASP knowledge modules we used for the experiments. 
Section~\ref{app:sec:dataset-errors} enumerates the errors in the datasets.

\NBB{Few shot and CoT prompts are given in the supplementary code file. ...}
For bAbI, the prompts for the baseline few-shot prompting can be found in the directory \path{bAbI_baseline/example_prompts}, while the prompts for chain-of-thought can be found in \path{bAbI_baseline/COT_prompts_v3}.
For StepGame, the prompts for the baseline few-shot prompting and chain-of-thought can be found in the directory \path{stepGame/prompts}. The following table records the cost for GPT-3 queries used in GPT-3 baselines and our method, where Eng. denotes the engine of GPT-3, c1, d2, d3 denote text-curie-001, text-davinci-002, and text-davinci-003.

\begin{center}
\small
\begin{tabular}{ |c|c|c|c|c| } 
 \hline
 Dataset & Method & Eng. & \#Data & Cost \\ 
 \hline
 \hline
 \multirow{3}{*}{bAbI} & Few-Shot & d3 & 20k & \$190 \\
  & CoT & d3 & 20k & \$280 \\ 
  & GPT-3+ASP & d3 & 20k & \$41 \\ 
 \hline
 \multirow{4}{*}{StepGame} & Few-Shot & d3 & 1k & \$21 \\
  & CoT & d3 & 1k & \$26 \\ 
  & GPT-3+ASP & c1 & 10k & \$89 \\ 
  & GPT-3+ASP & d2 & 10k & \$886 \\ 
 \hline
 CLUTRR 1.0 & GPT-3+ASP & d3 & 879 & \$37 \\
 \hline
 CLUTRR 1.3 & GPT-3+ASP & d3 & 400 & \$17 \\
 \hline
 CLUTRR-S & GPT-3+ASP & d3 & 563 & \$19 \\ 
 \hline
 gSCAN & GPT-3+ASP & c1 & 8k & \$0.2 \\ 
 \hline
 \multirow{2}{*}{Pick\&Place} & Few-Shot & d3 & 40 & \$0.5 \\ 
  & GPT-3+ASP & d3 & 40 & \$0.4 \\ 
 \hline
\end{tabular}
\end{center}

All experiments were conducted on Ubuntu 18.04.2 LTS with two 10-core CPU Intel(R) Xeon(R) CPU E5-2640 v4 @ 2.40GHz and four GP104 [GeForce GTX 1080] graphics cards.

All datasets used in this paper are publicly available.
The bAbI dataset is under BSD license.
The CLUTRR dataset is released under ``Attribution-NonCommercial 4.0 International'' license.
The StepGame dataset doesn't have a specified license.
The gSCAN dataset is released under MIT license.

\section{Robot Planning} \label{app:sec:robot}


Recently, there has been increasing interest in using LLMs to find a sequence of executable actions for robots, aiming to achieve high-level goals expressed in natural language, such as 
SayCan \cite{ahn22can} and Innermonologue \cite{huang22inner}. However, it is worth noting that the actions generated by LLMs tend to be loosely connected and do not take into account the intermediate state changes that occur during the execution of these actions. 

\begin{figure}[ht]
\begin{center}
\includegraphics[width=1.0\columnwidth]{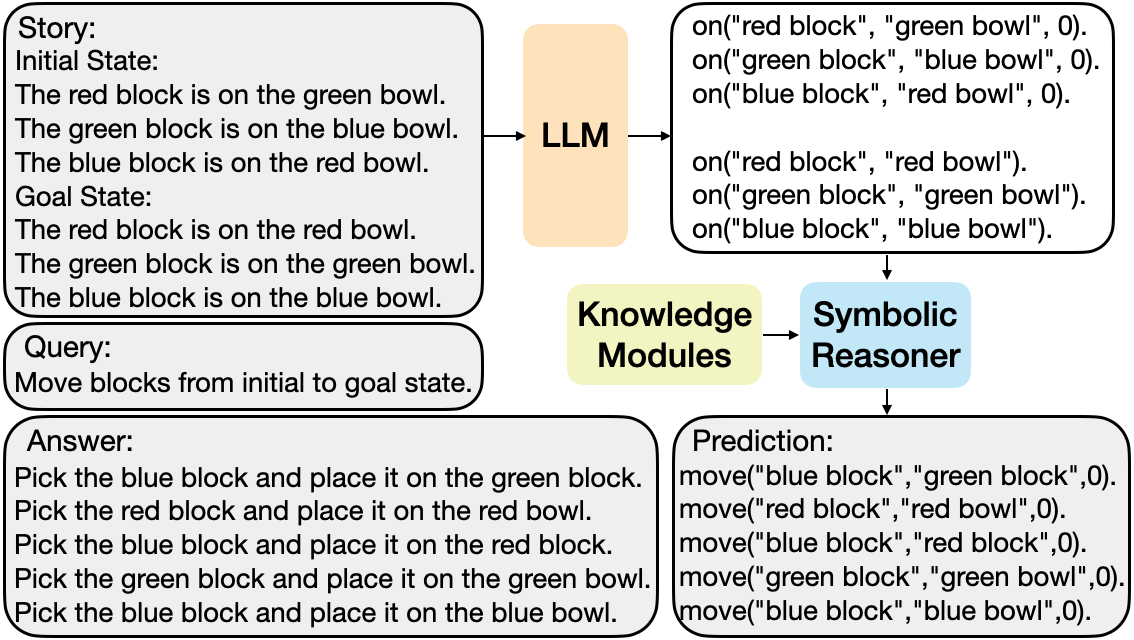}
\end{center}
\caption{The GPT-3+ASP pipeline for Pick\&Place}
\label{fig:pipeline_blockworld}
\end{figure}


We based our work on SayCan's open-source virtual tabletop environment\footnote{\url{https://github.com/google-research/google-research/tree/master/saycan}}, where a robot is tasked with achieving a goal, such as "stack the blocks," on a table with colored blocks and bowls. We noticed that the successful plans demonstrated by SayCan are restricted to simple one-step look-ahead plans that do not take into account intermediate state changes.
 
We randomly sampled 40 data instances of the form $\langle S_i, S_g, L\rangle$ in the Pick\&Place domain with 4 to 7 blocks and 3 to 7 bowls, possibly stacked together and with 3 to 10 steps of pick\_and\_place actions required by the robot to change the initial state $S_i$ to the goal state $S_g$.
Here, the label $L$ is the set of instructions to achieve the goals (e.g., ``1. Move the violet block onto the blue block. 2...''). Among 40 data instances, 20 data instances contain only blocks that can be placed on the table while 20 data instances contain both blocks and bowls and assume all blocks must be on the bowls.

The baseline for this dataset follows the method in SayCan's open-source virtual tabletop environment, where GPT-3 is used as the large language model to directly find the sequence of actions from $S_i$ to $S_g$. However, the baseline fails to find successful plans for all 40 randomly sampled data instances. This result confirms the claim by \cite{valmeekam22large} that large language models are not suitable as planners. 

We also applied our method to this task. We let GPT-3 turn the states $S_i$ and $S_g$ into atomic facts of the form $on(A,B,0)$ and $on(A,B)$, respectively. Then, an ASP program for the Pick\&Place domain is used to find an optimal plan. We found that while GPT-3 has only 0\% accuracy in predicting the whole plan, it has 100\% accuracy in fact extraction under the provided format. When we apply symbolic reasoning to these extracted atomic facts with an ASP program, we could achieve 100\% accuracy on the predicted plans.
Details of the prompts are available in Appendix~\ref{appendix:prompt:blockworld}.

\begin{table}[ht!]
	\centering
	{\small 
	\begin{tabular}{l|cc}
		\toprule
		 Method & Blocks & Blocks+Bowls \\
        \midrule
        GPT-3(d3) & 0 & 0 \\
        GPT-3(d3)+ASP & {\bf 100} & {\bf 100} \\
		\bottomrule
	\end{tabular}
	}
	\caption{Test accuracy on the Pick\&Place dataset. (d3) denotes the text-davinci-003 model.}
	\label{tb:blockworld}
\end{table}

\begin{figure}[ht]
\begin{center}
\includegraphics[width=1.0\columnwidth]{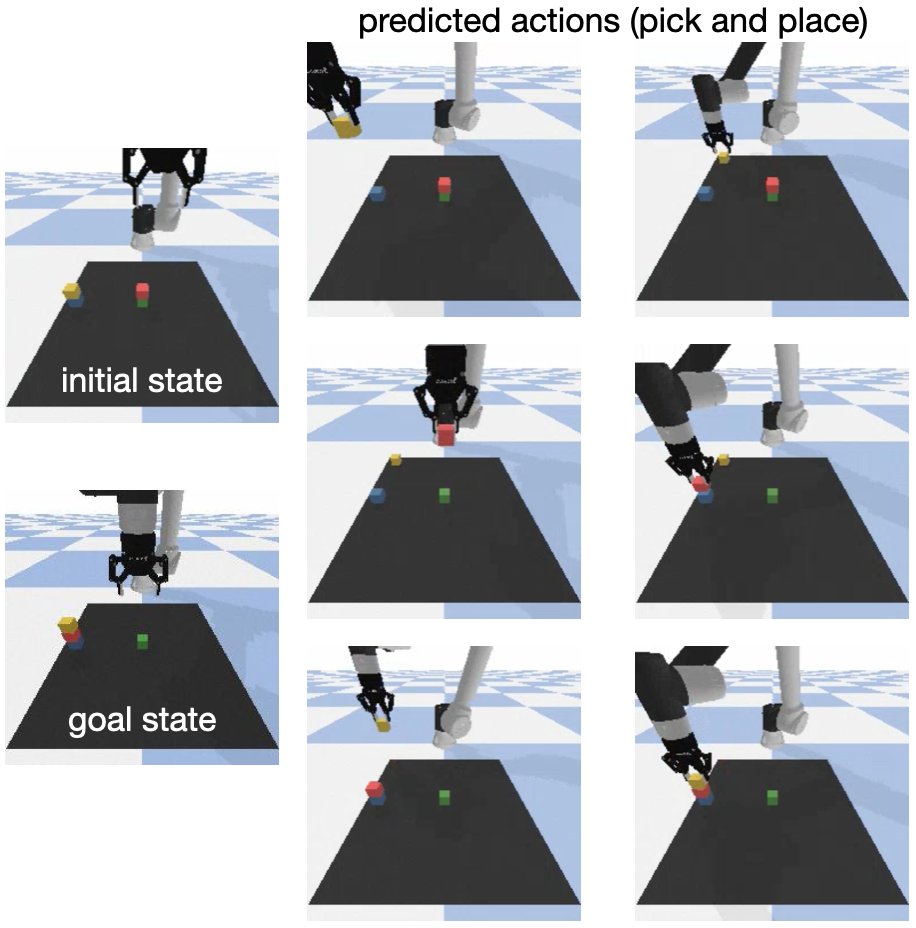}
\end{center}
\caption{A simple plan predicted by GPT-3+ASP in the Pick\&Place domain.}
\label{fig:saycan_example}
\end{figure}

\section{More about CLUTRR} \label{app:sec:clutrr}
\subsection{CLUTRR 1.3 Data Generation}\label{app:subsec:clutrr_gen}

We used CLUTRR 1.3 codes to generate $400$ test data instances.~\footnote{
We used the development branch of CLUTRR repository https://github.com/facebookresearch/clutrr/tree/develop.
}
Our generated CLUTRR 1.3 dataset consists of $100$ data for each of the four categories: (assuming that the query is asking about the relation between persons $A$ and $D$)
\begin{itemize}
\item clean: each story describes 3 relations in a chain of four persons $A-B-C-D$;
\item supporting: each story describes 3 relations in a chain of four persons $A-B-C-D$ as well as an additional relation $X-Y$ such that $X,Y\in \{A,B,C,D\}$ and $X-Y$ is not the queried pair;
\item irrelevant: each story describes 3 relations in a chain of four persons $A-B-C-D$ as well as an additional relation $X-Y$ such that $X\in \{A,B,C,D\}$ and $Y\not \in \{A,B,C,D\}$;
\item disconnected: each story describes 3 relations in a chain of four persons $A-B-C-D$ as well as an additional relation $X-Y$ such that $X,Y\not \in \{A,B,C,D\}$.
\end{itemize}

\subsection{Evaluation on CLUTRR 1.0}\label{appendix:dataset:clutrr:eval}

\begin{table}[ht!]
	\centering
	{\small 
	\begin{tabular}{cc|cccc}
		\toprule
        Training & Testing & BA & GSM & d2 & d3\\
        \hline
        \multirow{4}{*}{Clean} & Clean & 58 & {\bf 69} & 63 & 68 \\
        & Supporting & {\bf 76} & 66 & 62 & 62 \\
        & Irrelevant & 70 & {\bf 77} & 66 & 71 \\
        & Disconnected & 49 & 36 & {\bf 59} & {\bf 59} \\
        \hline
        Supporting & Supporting & 67 & 49 & {\bf 83} & {\bf 83} \\
        \hline
        Irrelevant & Irrelevant & 51 & 63 & 72 & {\bf 75} \\
        \hline
        Disconnected & Disconnected & 57 & 53 & 63 & {\bf 67} \\
 	\bottomrule
	\end{tabular}
	}
	\caption{Test accuracy on the CLUTRR dataset. BA denotes BiLSTM-Attention. d2 and d3 denote GPT-3+ASP with text-davinci-002 and text-davinci-003 model.}
	\label{tb:clutrr}
\end{table}

Table~\ref{tb:clutrr} compares the accuracy of our method and the state-of-the-art methods, i.e., BiLSTM-Attention \citep{sinha19clutrr} and GSM (with a BiLSTM encoder) \citep{tian21generative}, on the (original) CLUTRR test dataset. Except for our method, all other models are trained on a specific split of the CLUTRR training dataset.

\begin{table}[ht!]
	\centering
	{\small 
	\begin{tabular}{cc|ccc}
		\toprule
        Training & Testing & DP & d2 & d3\\
        \hline
        \multirow{4}{*}{Clean} & Clean & {\bf 100} & {\bf 100} & {\bf 100} \\
        & Supporting & {\bf 99} & 96 & {\bf 99} \\
        & Irrelevant & 98 & 99 & {\bf 100} \\
        & Disconnected & 99 & 98 & {\bf 100} \\
        \hline
        Supporting & Supporting & {\bf 100} & {\bf 100} & {\bf 100} \\
        \hline
        Irrelevant & Irrelevant & {\bf 100} & 97 & {\bf 100} \\
        \hline
        Disconnected & Disconnected & 94 & 97 & {\bf 100} \\
 	\bottomrule
	\end{tabular}
	}
	\caption{Test accuracy on the CLUTRR-S dataset. DP denotes DeepProbLog, d2 and d3 denote GPT-3+ASP with the text-davinci-002 and text-davinci-003 model.}
	\label{tb:clutrr-s:app}
\end{table}

Table~\ref{tb:clutrr-s:app} compares the accuracy of our method and the state-of-the-art method,  DeepProbLog \citep{manhaeve21neural} on the CLUTRR-S test dataset. With GPT-3(d2)+ASP on the CLUTRR-S dataset, 550 out of 563 predictions are correct, and 13 are wrong. All errors occur due to the entities in a relation being swapped. For example, we use ``{\tt son(A,B)}'' to represent ``A has a son B'' while GPT-3 text-davinci-002 responded with ``{\tt  son(Robert,Ryan)}'' for the sentence ``Robert is Ryan's son.'' On the other hand, text-davinci-003 performed better, with only a single error and 562 out of 563 predictions being correct.

\section{Prompts for Semantic Parsing}\label{app:sec:prompts}

Below, we present the details of the general knowledge of the prompts that we summarized and applied in this paper, followed by some examples.
\begin{enumerate}
\item If the information in a story (or query) can be extracted independently, parsing each sentence separately (using the same prompt multiple times) typically works better than parsing the whole story. Since people usually cache all GPT-3 responses to save cost by avoiding duplicated GPT-3 requests for the same prompt, parsing each sentence separately also yields better usage of cached responses. Below are some examples.

\begin{itemize}
    \item In most bAbI tasks (except for tasks 11 and 13), the sentences in a story (including the query sentence) are independent of each other. We parse each sentence separately using GPT-3 as in the Appendix~\ref{appendix:ssec:bAbI}.
    \item In the stepGame dataset, each sentence in a story describes the spatial relation between 2 objects. There are 4 sentences in a story when $k=1$ and about 20 sentences when $k=10$. If we ask GPT-3 to extract all the atomic facts from the whole story, it always misses some atoms or predicts wrong atoms. Since every sentence is independent of each other as shown in Figure~\ref{fig:pipeline}, we use the following (truncated) prompt multiple times for each data instance where each time {\tt [INPUT]} is replaced with one sentence in the story or the query. This yields a much higher accuracy as in Section~\ref{ssec:CLUTRR-experiment}. The complete prompt is available in Appendix~\ref{appendix:prompt:stepgame}.
\begin{lstlisting}
Please parse each sentence into a fact. If the sentence is describing clock-wise information, then 12 denotes top, 1 and 2 denote top_right, 3 denotes right, ... If the sentence is describing cardinal directions, then north denotes top, ...

Sentence: What is the relation of the agent X to the agent K?
Semantic Parse: query("X", "K").

Sentence: H is positioned in the front right corner of M.
Semantic Parse: top_right("H", "M").
...
Sentence: [INPUT]
Semantic Parse:
\end{lstlisting}
\end{itemize}
However, if some sentences in a story are dependent, splitting them may lead to unexpected results in the GPT-3 response. Below are some examples.
\begin{itemize}
    \item In bAbI task \#11 and  \#13, a story may contain the two consecutive sentences ``Mary went back to the bathroom.
After that she went to the bedroom.'' There is a dependency on the sentences to understand that ``she'' in the second sentence refers to ``Mary'' in the first. For this reason, task \#11 stories are parsed as a whole. This is similar for task \#13. 
    \item In the CLUTRR dataset, a story may contain sentences with coreferences like ``Shirley enjoys playing cards with her brother. His name is Henry.'' where the latter sentence depends on the former one, and a family relation can be correctly extracted only with both sentences. Thus for CLUTRR datasets (i.e., CLUTRR 1.0, CLUTRR 1.3, and CLUTRR-S), we extract the family relations and gender relations from the whole story.
\end{itemize}

\item  There is certain commonsense knowledge that GPT-3 is not aware of, and describing the missing knowledge in the prompt works better than adding examples only. This happens when GPT-3 cannot generalize such knowledge well with a few examples.
\begin{itemize}
    \item For example, in StepGame dataset, clock numbers are used to denote cardinal directions, e.g., ``H is below J at 4 o'clock'' means ``H is on the bottom-right of J''. Such knowledge in the dataset is not well captured by GPT-3 and enumerating examples in the prompt doesn't work well. On the other hand, describing such knowledge at the beginning of the prompt as shown in Appendix~\ref{appendix:prompt:stepgame} increases the accuracy by a large margin.
\end{itemize}

\end{enumerate}

\subsection{bAbI}\label{appendix:ssec:bAbI}

For bAbI dataset, there are two prompts for each task, corresponding to the context and query. Each prompt has a consistent set of basic instructions followed by example pairs of text and parsed text. 
Below are the prompts used to parse the context and query facts from a story and query, where {\tt [Input]} at the end is replaced with the story in each test data instance.
We only present the prompts for Tasks 1,2, and 3. The rest of 
the prompts can be found in the repository in \url{https://github.com/azreasoners/LLM-ASP/blob/main/bAbI/GPT_prompts.py}.

\medskip
\noindent{\bf Tasks 1/2/3 (Context)}

\begin{lstlisting}
Please parse the following statements into facts. The available keywords are: pickup, drop, and go.
Sentence: Max journeyed to the bathroom.
Semantic parse: go(Max, bathroom).

Sentence: Mary grabbed the football there.
Semantic parse: pickup(Mary, football).

Sentence: Bob picked up the apple.
Semantic parse: pickup(Bob, apple).

Sentence: Susan dropped the milk.
Semantic parse: drop(Susan, milk).

Sentence: Bob got the football there.
Semantic parse: pickup(Bob, football).

Sentence: Max left the cup.
Semantic parse: drop(Max, cup).

Sentence: Kevin put down the pie there.
Semantic parse: drop(Kevin, pie).

Sentence: John took the football there.
Semantic parse: pickup(John, football).

Sentence: [INPUT]
Semantic parse:
\end{lstlisting}

\noindent{\bf Task 1 (Query)}

\begin{lstlisting}
Please parse the following questions into query facts. The available keywords are: whereAgent.
Sentence: Where is Mary?
Semantic parse: whereAgent(Mary).

Sentence: Where is Daniel?
Semantic parse: whereAgent(Daniel).

Sentence: Where is Sandra?
Semantic parse: whereAgent(Sandra).

Sentence: Where is John?
Semantic parse: whereAgent(John).

Sentence: [INPUT]
Semantic parse:
\end{lstlisting}

\noindent{\bf Task 2 (Query)}

\begin{lstlisting}
Please parse the following questions into query facts. The available keywords are: loc.
Sentence: Where is the toothbrush?
Semantic parse: loc(toothbrush).

Sentence: Where is the milk?
Semantic parse: loc(milk).

Sentence: Where is the apple?
Semantic parse: loc(apple).

Sentence: Where is the football?
Semantic parse: loc(football).

Sentence: [INPUT]
Semantic parse:
\end{lstlisting}

\noindent{\bf Task 3 (Query)}

\begin{lstlisting}
Please parse the following questions into query facts. The available keywords are: loc.
Sentence: Where was the football before the bathroom?
Semantic parse: before(football,bathroom).

Sentence: Where was the apple before the garden?
Semantic parse: before(apple,garden).

Sentence: Where was the milk before the kitchen?
Semantic parse: before(milk,kitchen).

Sentence: Where was the apple before the bedroom?
Semantic parse: before(apple,bedroom).

Sentence: Where was the football before the hallway?
Semantic parse: before(football,hallway).

Sentence: [INPUT]
Semantic parse:
\end{lstlisting}

\subsection{StepGame}\label{appendix:prompt:stepgame}

For the StepGame dataset, there is only one prompt below to extract the location relations among objects. All example sentences are from the training data in (the noise split of) the original StepGame dataset.\footnote{
\url{https://github.com/ZhengxiangShi/StepGame/tree/main/Code/babi_format/noise}
}
The {\tt [Input]} at the end of the prompt is replaced with each sentence in a test data instance.
\begin{lstlisting}
Please parse each sentence into a fact. If the sentence is describing clock-wise information, then 12 denotes top, 1 and 2 denote top_right, 3 denotes right, 4 and 5 denote down_right, 6 denotes down, 7 and 8 denote down_left, 9 denote left, 10 and 11 denote top_left. If the sentence is describing cardinal directions, then north denotes top, east denotes right, south denotes down, and west denotes left. If the sentence is a question, the fact starts with query. Otherwise, the fact starts with one of top, down, left, right, top_left, top_right, down_left, and down_right. 

Sentence: What is the relation of the agent X to the agent K?
Semantic Parse: query("X", "K").

Sentence: H is positioned in the front right corner of M.
Semantic Parse: top_right("H", "M").

Sentence: F is on the left side of and below Q.
Semantic Parse: down_left("F", "Q").

Sentence: Y and I are parallel, and Y is on top of I.
Semantic Parse: top("Y", "I").

Sentence: V is over there with T above.
Semantic Parse: top("T", "V").

Sentence: V is slightly off center to the top left and G is slightly off center to the bottom right.
Semantic Parse: top_left("V", "G").

Sentence: The objects S and A are over there. The object S is lower and slightly to the left of the object A.
Semantic Parse: down_left("S", "A").

Sentence: D is diagonally below Z to the right at a 45 degree angle.
Semantic Parse: down_right("D", "Z").

Sentence: V is at A's 9 o'clock.
Semantic Parse: left("V", "A").

Sentence: J is at O's 6 o'clock.
Semantic Parse: down("J", "O").

Sentence: H is below J at 4 o'clock.
Semantic Parse: down_right("H", "J").

Sentence: O is there and C is at the 5 position of a clock face.
Semantic Parse: down_right("C", "O").

Sentence: If H is the center of a clock face, B is located between 10 and 11.
Semantic Parse: top_left("B", "H").

Sentence: [Input]
Semantic Parse:
\end{lstlisting}


\subsection{CLUTRR}\label{appendix:prompt:clutrr}
For CLUTRR dataset, there are two prompts to extract the family relations and genders from a story respectively. All example stories in both prompts are from the training data ``data\_06b8f2a1\slash 2.2,2.3\_train.csv'' in the original CLUTRR dataset.%
\footnote{The original CLUTRR data is available in \url{https://github.com/facebookresearch/clutrr}.}
Below is the prompt to extract family relations from a story where {\tt [Input]} at the end is replaced with the story in each test data instance.
\begin{lstlisting}
Given a story, extract atomic facts of the form relation("Person", "Person"). Example relations are: father, mother, parent, son, daughter, child, grandfather, grandmother, grandson, granddaughter, wife, husband, spouse, sibling, nephew, niece, uncle, aunt, child_in_law, and parent_in_law.

Story: [Verdie] waved good bye to her dad [Henry] for the day and went next door with her sister [Amanda]. [Henry]'s daughter, [Amanda], went to the city this weekend. She spent her time there visiting her grandfather, [Kyle], and had a wonderful time with him.
Semantic Parse: father("Verdie", "Henry"). sister("Verdie", "Amanda"). daughter("Henry", "Amanda"). grandfather("Amanda", "Kyle").

Story: [Michelle] was excited for today, its her daughter's, [Theresa], spring break. She will finally get to see her. [Michael] was busy and sent his wife, [Marlene], instead. [Kristen] loved to care for her newborn child [Ronald]. [Eric]'s son is [Arthur].
Semantic Parse: daughter("Michelle", "Theresa"). wife("Michael", "Marlene"). child("Kristen", "Ronald"). son("Eric", "Arthur").

Story: [Vernon] was present in the delivery room when his daughter [Raquel] was born, but when his daughter [Constance] was born he was too sick. [Vernon] and his daughter [Margaret] went to the movies. [Constance], [Margaret]'s sister, had to stay home as she was sick.
Semantic Parse: daughter("Vernon", "Raquel"). daughter("Vernon", "Constance"). daughter("Vernon", "Margaret"). sister("Margaret", "Constance").

Story: [Eric] who is [Carl]'s father grounded [Carl] after finding out what [Carl] had done at school. [Ronald] was busy planning a 90th birthday party for his aunt, [Theresa]. [Eric] and his son [Carl] went to the park and saw [Eric]'s father [Kyle] there with his dog.
Semantic Parse: father("Carl", "Eric"). aunt("Ronald", "Theresa"). son("Eric", "Carl"). father("Eric", "Kyle").

Story: [Shirley] and [Edward] are siblings and best friends. They do everything together. [Henry] walked his daughters [Amanda] and [Michelle] to school. [Kyle] enjoys watching movies with his son's daughter. Her name is [Amanda].
Semantic Parse: sibling("Shirley", "Edward"). daughter("Henry", "Amanda"). daughter("Henry", "Michelle"). granddaughter("Kyle", "Amanda").

Story: [Raquel] and her brother [Casey] took her grandmother [Karen] to the store to buy a new dress. [Karen] and her husband [Kyle] just celebrated 10 years of marriage. [Karen] loves her grandson, [Casey], and he loves her too.
Semantic Parse: brother("Raquel", "Casey"). grandmother("Raquel", "Karen"). husband("Karen", "Kyle"). grandson("Karen", "Casey").

Story: [Allen]'s father, [Eric], bought him some ice cream. [Karen] was baking cookies for her grandson, [Allen]. [Allen]'s brother [Arthur] came home from school, so she baked some extra for him, too. [Eric]'s son, [Arthur], was ill and needed to be picked up at school. [Eric] hurried to his side.
Semantic Parse: father("Allen", "Eric"). grandson("Karen", "Allen"). brother("Allen", "Arthur"). son("Eric", "Arthur").

Story: [Karen] was spending the weekend with her grandson, [Eddie]. [Eddie]'s sister [Michelle] was supposed to come too, but she was busy and could n't make it. [Theresa] took her daughter, [Michelle], out to High Tea yesterday afternoon. [Eddie]'s mother [Theresa] baked brownies for dessert after they had dinner.
Semantic Parse: grandson("Karen", "Eddie"). sister("Eddie", "Michelle"). daughter("Theresa", "Michelle"). mother("Eddie", "Theresa").

Story: [Input]
Semantic Parse:
\end{lstlisting}


We also use a variant of the above prompt to extract the gender of each person in a story. The prompt context is a bit simpler as there are only two genders. The examples are the same while the {\tt Semantic Parse} result is simply replaced with the atomic facts about gender information. 
Below is the prompt to extract the gender of each person in a story where {\tt [Input]} is replaced with the story in each test data instance.
\begin{lstlisting}
Given a story, extract atomic facts of the form male("Person") or female("Person") for every person that appears in the sentences.

Story: [Verdie] waved good bye to her dad [Henry] for the day and went next door with her sister [Amanda]. [Henry]'s daughter, [Amanda], went to the city this weekend. She spent her time there visiting her grandfather, [Kyle], and had a wonderful time with him.
Semantic Parse: female("Verdie"). male("Henry"). female("Amanda"). male("Kyle").

Story: [Michelle] was excited for today, its her daughter's, [Theresa], spring break. She will finally get to see her. [Michael] was busy and sent his wife, [Marlene], instead. [Kristen] loved to care for her newborn child [Ronald]. [Eric]'s son is [Arthur].
Semantic Parse: female("Michelle"). female("Theresa"). male("Michael"). female("Marlene"). female("Kristen"). male("Ronald"). male("Eric"). male("Arthur").

Story: [Vernon] was present in the delivery room when his daughter [Raquel] was born, but when his daughter [Constance] was born he was too sick. [Vernon] and his daughter [Margaret] went to the movies. [Constance], [Margaret]'s sister, had to stay home as she was sick.
Semantic Parse: male("Vernon"). female("Raquel"). female("Constance"). female("Margaret").

Story: [Eric] who is [Carl]'s father grounded [Carl] after finding out what [Carl] had done at school. [Ronald] was busy planning a 90th birthday party for his aunt, [Theresa]. [Eric] and his son [Carl] went to the park and saw [Eric]'s father [Kyle] there with his dog.
Semantic Parse: male("Eric"). male("Carl"). male("Ronald"). female("Theresa"). male("Kyle").

Story: [Shirley] and [Edward] are siblings and best friends. They do everything together. [Henry] walked his daughters [Amanda] and [Michelle] to school. [Kyle] enjoys watching movies with his son's daughter. Her name is [Amanda].
Semantic Parse: female("Shirley"). male("Edward"). male("Henry"). female("Amanda"). female("Michelle"). male("Kyle").

Story: [Raquel] and her brother [Casey] took her grandmother [Karen] to the store to buy a new dress. [Karen] and her husband [Kyle] just celebrated 10 years of marriage. [Karen] loves her grandson, [Casey], and he loves her too.
Semantic Parse: female("Raquel"). male("Casey"). female("Karen"). male("Kyle").

Story: [Allen]'s father, [Eric], bought him some ice cream. [Karen] was baking cookies for her grandson, [Allen]. [Allen]'s brother [Arthur] came home from school, so she baked some extra for him, too. [Eric]'s son, [Arthur], was ill and needed to be picked up at school. [Eric] hurried to his side.
Semantic Parse: male("Allen"). male("Eric"). female("Karen"). male("Arthur").

Story: [Karen] was spending the weekend with her grandson, [Eddie]. [Eddie]'s sister [Michelle] was supposed to come too, but she was busy and could n't make it. [Theresa] took her daughter, [Michelle], out to High Tea yesterday afternoon. [Eddie]'s mother [Theresa] baked brownies for dessert after they had dinner.
Semantic Parse: female("Karen"). male("Eddie"). female("Michelle"). female("Theresa").

Story: [Input]
Semantic Parse:
\end{lstlisting}

For CLUTRR-S dataset, i.e., the simpler version of the CLUTRR dataset from DeepProbLog \citep{manhaeve21neural} repository, there are also two prompts below to extract the family relations and genders from a story respectively.\footnote{
The CLUTRR-S dataset is from \url{https://github.com/ML-KULeuven/deepproblog/tree/master/src/deepproblog/examples/CLUTRR/data}.
}
All example stories in both prompts are from the training data ``data\_a7d9402e\slash 2.2,2.3\_train.csv''.
\begin{lstlisting}
Given a story, extract atomic facts of the form relation("Person", "Person") about family relationships that appear in the sentences.

Story: [Mervin] is [Robert]'s father.  [Robert] is the father of [Jim].  [Jon] is [Robert]'s brother.  [Mervin] is the father of [Jon].
Semantic Parse: father("Robert", "Mervin"). father("Jim", "Robert"). brother("Robert", "Jon"). father("Jon", "Mervin").

Story: [Brooke] is [Cheryl]'s sister.  [Jon] is the father of [Brooke].  [Melissa] is [Jon]'s mother.  [Jon] is [Cheryl]'s father.
Semantic Parse: sister("Cheryl", "Brooke"). father("Brooke", "Jon"). mother("Jon", "Melissa"). father("Cheryl", "Jon").

Story: [Jon] is [Carol]'s brother.  [Carol] is [Joyce]'s mother.  [Helen] is [Carol]'s sister.  [Helen] is a sister of [Jon].
Semantic Parse: brother("Carol", "Jon"). mother("Joyce", "Carol"). sister("Carol", "Helen"). sister("Jon", "Helen").

Story: [Melissa] is [Glenn]'s grandmother.  [Melissa] is the mother of [Calvin].  [Glenn] is a son of [Lila].  [Calvin] is [Glenn]'s father.
Semantic Parse: grandmother("Glenn", "Melissa"). mother("Calvin", "Melissa"). son("Lila", "Glenn"). father("Glenn", "Calvin").

Story: [Margaret] has a brother named [William].  [William] is [Carol]'s son.  [Margaret] is [Carol]'s daughter.  [Lila] is the aunt of [William].
Semantic Parse: brother("Margaret", "William"). son("Carol", "William"). daughter("Carol", "Margaret"). aunt("William", "Lila").

Story: [Stephanie] is a sister of [Lois].  [Lois] is [Theresa]'s sister.  [Helen] is [Lois]'s mother.  [Helen] is [Stephanie]'s mother.
Semantic Parse: sister("Lois", "Stephanie"). sister("Theresa", "Lois"). mother("Lois", "Helen"). mother("Stephanie", "Helen").

Story: [Jon] is [Elias]'s brother.  [Michael] is a son of [Helen].  [Jon] is the uncle of [Michael].  [Elias] is the father of [Michael].
Semantic Parse: brother("Elias", "Jon"). son("Helen", "Michael"). uncle("Michael", "Jon"). father("Michael", "Elias").

Story: [Carol] has a son called [William].  [Melissa] is the mother of [Jon].  [Jon] is the uncle of [William].  [Carol] has a brother named [Jon].
Semantic Parse: son("Carol", "William"). mother("Jon", "Melissa"). uncle("William", "Jon"). brother("Carol", "Jon").

Story: [Robert] is the father of [Jim].  [Robert] has a daughter called [Ashley].  [Elias] is [Robert]'s brother.  [Elias] is the uncle of [Ashley].
Semantic Parse: father("Jim", "Robert"). daughter("Robert", "Ashley"). brother("Robert", "Elias"). uncle("Ashley", "Elias").

Story: [Elias] is the father of [Carlos].  [Elias] is the father of [Andrew].  [Andrew] is [Carlos]'s brother.  [Jon] is a brother of [Elias].
Semantic Parse: father("Carlos", "Elias"). father("Andrew", "Elias"). brother("Carlos", "Andrew"). brother("Elias", "Jon").

Story: [Jon] is the father of [Ben].  [James] is [Kevin]'s brother.  [Ben] is a brother of [James].  [Jon] is [James]'s father.
Semantic Parse: father("Ben", "Jon"). brother("Kevin", "James"). brother("James", "Ben"). father("James", "Jon").

Story: [Carol] has a sister named [Lila].  [William] is [Carol]'s son.  [Helen] is [Lila]'s sister.  [Lila] is [William]'s aunt.
Semantic Parse: sister("Carol", "Lila"). son("Carol", "William"). sister("Lila", "Helen"). aunt("William", "Lila").

Story: [Calvin] is [Bruce]'s father.  [Elias] is [Calvin]'s brother.  [Calvin] is [Kira]'s father.  [Kira] is [Bruce]'s sister.
Semantic Parse: father("Bruce", "Calvin"). brother("Calvin", "Elias"). father("Kira", "Calvin"). sister("Bruce", "Kira").

Story: [Carol] is a sister of [Helen].  [Carol] is [Carlos]'s aunt.  [Lila] is [Carol]'s sister.  [Carlos] is [Helen]'s son.
Semantic Parse: sister("Helen", "Carol"). aunt("Carlos", "Carol"). sister("Carol", "Lila"). son("Helen", "Carlos").

Story: [Input]
Semantic Parse:
\end{lstlisting}
Note that, although the sentences in the CLUTRR-S dataset is much simpler than those in CLUTRR dataset, we don't achieve 100\% accuracy in GPT-3 responses with the above long prompt. This is partially because the above prompt violates prompting strategy 3 in Section~\ref{sec:method} as the order of names in a binary relation in sentences is mostly following ``{\tt relationOf(A,B)}'' instead of ``{\tt relation(B,A)}''.

\begin{lstlisting}
Given a story, extract atomic facts of the form male("Person") or female("Person") for every person that appears in the sentences.

Story: [Jon] is [Carol]'s brother.  [Mervin] has a daughter called [Carol].  [Chantell] is a daughter of [Jon].  [Mervin] has a son called [Jon].
Semantic Parse: male("Jon"). female("Carol"). male("Mervin"). female("Chantell").

Story: [Melissa] is [Glenn]'s grandmother.  [Melissa] is the mother of [Calvin].  [Glenn] is a son of [Lila].  [Calvin] is [Glenn]'s father.
Semantic Parse: female("Melissa"). male("Glenn"). male("Calvin"). female("Lila").

Story: [Input]
Semantic Parse:
\end{lstlisting}

\subsection{gSCAN}\label{appendix:prompt:gscan}
For gSCAN dataset, there is only one prompt below to extract the command in each data instance. All example sequences are from the training data.\footnote{
\url{https://github.com/LauraRuis/groundedSCAN/tree/master/data/compositional_splits.zip}
}
The {\tt [Input]} at the end of the prompt is replaced with the command in each test data instance.
\begin{lstlisting}
Please parse each sequence of words into facts. 

Sequence: pull a yellow small circle
Semantic Parse: query(pull). queryDesc(yellow). queryDesc(small). queryDesc(circle).

Sequence: push a big square
Semantic Parse: query(push). queryDesc(big). queryDesc(square).

Sequence: push a green small square cautiously
Semantic Parse: query(push). queryDesc(green). queryDesc(small). queryDesc(square). while(cautiously).

Sequence: pull a circle hesitantly
Semantic Parse: query(pull). queryDesc(circle). while(hesitantly).

Sequence: walk to a yellow big cylinder while spinning
Semantic Parse: query(walk). queryDesc(yellow). queryDesc(big). queryDesc(cylinder). while(spinning).

Sequence: push a big square while zigzagging
Semantic Parse: query(push). queryDesc(big). queryDesc(square). while(zigzagging).

Sequence: push a cylinder hesitantly
Semantic Parse: query(push). queryDesc(cylinder). while(hesitantly).

Sequence: [Input]
Semantic Parse:
\end{lstlisting}

\subsection{Pick\&Place}\label{appendix:prompt:blockworld}
For the Pick\&Place dataset, there are two prompts below to extract the atomic facts from the initial state and the goal state, respectively.
\begin{lstlisting}
Turn each sentence into an atomic fact of the form on(A, B, 0).

Sentence: The red block is on the yellow bowl.
Semantic Parse: on("red block", "yellow bowl", 0).

Sentence: The violet block is on the blue block.
Semantic Parse: on("violet block", "blue block", 0).

Sentence: [INPUT]
Semantic Parse:
\end{lstlisting}
\begin{lstlisting}
Turn each sentence into an atomic fact of the form on(A, B).

Sentence: The red block is on the yellow bowl.
Semantic Parse: on("red block", "yellow bowl").

Sentence: The violet block is on the blue block.
Semantic Parse: on("violet block", "blue block").

Sentence: [INPUT]
Semantic Parse:
\end{lstlisting}
For each sentence in the initial or goal state, we replace {\tt [INPUT]} in the corresponding prompt above with this sentence and request GPT-3 to extract a single atomic fact. The union of these atomic facts extracted from all sentences is then used in the symbolic reasoner module to find an optimal plan.

For the GPT-3 baseline, we use the following prompt to let GPT-3 directly find a plan where {\tt [INPUT]} at the end of the prompt is replaced with the initial and goal state of the queried data instance.
\begin{lstlisting}
Find a shortest plan to move blocks from an initial state to a goal state. Note that you cannot move a block if anything is on it. You cannot move a block onto a target block or bowl if there is anything is on the target block or bowl. At most two blocks can be placed in the same bowl with one on top of the other.

# Initial State:
Nothing is on the green bowl.
The violet block is on the blue bowl.
The blue block is on the violet bowl.
The green block is on the blue block.

# Goal State:
The violet block is on the green bowl.
The green block is on the violet block.
The blue block is on the blue bowl.
Nothing is on the violet bowl.

Plan:
1. Move the violet block onto the green bowl.
2. Move the green block onto the violet block.
3. Move the blue block onto the blue bowl.

# Initial State:
Nothing is on the blue bowl.
The yellow block is on the green bowl.
The green block is on the violet bowl.
The violet block is on the green block.
The blue block is on the yellow bowl.
The red block is on the blue block.

# Goal State:
The yellow block is on the blue bowl.
The green block is on the yellow block.
The red block is on the green bowl.
Nothing is on the violet bowl.
The blue block is on the yellow bowl.
The violet block is on the blue block.

Plan:
1. Move the yellow block onto the blue bowl.
2. Move the red block onto the green bowl.
3. Move the violet block onto the blue block.
4. Move the green block onto the yellow block.

[INPUT]

Plan:
\end{lstlisting}



\section{GPT-3 Errors in Semantic Parsing}\label{app:sec:gpt3-errors}

In this section, we group and record the errors in the GPT-3 responses in tables where 
each row records a 3-tuple $\langle$ dataset, sentence(s), GPT-3 response $\rangle$. In this section, we list the following. 
\begin{itemize}
\item all 21 errors for the CLUTRR 1.3 dataset with text-davinci-003;
\item the single mistake in the first 100 data instances for every $k\in \{1,\dots,10\}$ in the StepGame dataset with text-davinci-002.
\end{itemize}

\subsection{Argument misorder}
A common mistake in the GPT-3 response is that the relation and arguments for an atom are correctly extracted, but the order of the arguments is incorrect. Such mistakes can be greatly alleviated by proper few-shot prompting where the orders of arguments in the example target atoms follow their orders in the stories.

\begin{center}
\small
\begin{tabular}{| m{12mm} | m{32mm} | m{23mm}| } 
\hline
CLUTRR 1.3 &  [Erasmus] bought to dress for his father [Joeseph] &  father("Joeseph", "Erasmus") \\
\hline
CLUTRR 1.3 & [Baldwin] bought to dress for his father [Sidney] & father("Sidney", "Baldwin") \\
\hline
CLUTRR 1.3 & [Elizabeth] is the uncle of [Leopold]. & uncle("Elizabeth", "Leopold") \\
\hline
\end{tabular}
\end{center}

There are only 3 errors in CLUTRR 1.3 due to argument misorder. The first 2 mistakes are indeed due to their missing periods at the end of the sentences -- if we simply add the periods back, their GPT-3 responses would become correct.

\subsection{Wrong relation}
Sometimes the arguments are correct, but the relations extracted by GPT-3 are incorrect or cannot be recognized by the ASP program. 

\begin{center}
\small
\begin{tabular}{| m{12mm} | m{32mm} | m{20mm}| } 
\hline
CLUTRR 1.3 & [Leila] and [Enoch] have been married for 35 years. & married("Leila", "Enoch") \\
\hline
CLUTRR 1.3 & [Lottie] and her grandsons [Clinton] and [Enoch] picked berries from the garden. & grandsons( "Lottie", "Clinton") grandsons( "Lottie", "Enoch") \\
\hline
CLUTRR 1.3 & [Felix] loved going to the store with his mom [Christena]. & mom("Felix", "Christena") \\
\hline
CLUTRR 1.3 & [Archibald] asked his father, [Abraham], to watch over his two sons, [Justus] and [Merle], for the weekend. & sons( "Archibald", "Justus", "Merle") \\
\hline
StepGame & If E is the center of a clock face, H is located between 2 and 3. & right("H", "E") \\
\hline
\end{tabular}
\end{center}

These kinds of mistake may be resolved by restricting the space of possible relations. 
For example, the mistakes in the first four rows can be resolved by simply adding the sentence ``Use spouse("Person", "Person") if two persons are couples.'' in the prompt.

\subsection{Ambiguious or incorrect co-reference}

\begin{center}
\small
\begin{tabular}{| m{12mm} | m{32mm} | m{23mm}| } 
\hline
CLUTRR 1.3 & [Katherine] was eating lunch with his son [Daniel] and his wife [Adelle]. & son("Katherine", "Daniel")
 wife("Daniel", "Adelle") \\
\hline
CLUTRR 1.3 & [Sylvester] and his son [Rollie] went to his mother [Ottilia]'s home & son("Sylvester", "Rollie") mother("Rollie", "Ottilia") \\
\hline
CLUTRR 1.3 & [Elsie] and [Lewis] did the Father daughter dance at the recital and his wife [Dollie] was so happy she cried &  father("Lewis", "Dollie") wife("Lewis", "Dollie") \\
\hline
\end{tabular}
\end{center}

\subsection{Anonymous argument}

\begin{center}
\small
\begin{tabular}{| m{12mm} | m{32mm} | m{23mm}| }
\hline
CLUTRR 1.3 & [Abe] is proud of his daughter's son. His name is [Johnie]. & son("Johnie", "Abe") \\
\hline
CLUTRR 1.3 & [Watt] and [Celestine] asked their mother, if they could go play in the pool. & mother("Watt", "Celestine") \\
\hline
CLUTRR 1.3 & [Davie] told his wife [Kitty] to pick up their daughter. & wife("Davie", "Kitty") daughter("Davie", "Kitty") \\
\hline
CLUTRR 1.3 & [Ethyl] just got married to her now husband, and in the process she got herself a daughter-in-law named [Henry]. & husband("Ethyl", "Michael").
 daughter\_in\_law( "Ethyl", "Henry") \\
 \hline
CLUTRR 1.3 & [Barnett] and [Manie] asked their mother, if they could go play in the pool. & mother("Barnett", "Manie") \\
\hline
CLUTRR 1.3 & [Santiago] told his wife [Octavia] to pick up their daughter. & wife("Santiago", "Octavia") daughter("Santiago", "Octavia") \\
\hline
\end{tabular}
\end{center}

\subsection{Missed to generate some atoms}

\begin{center}
\small
\begin{tabular}{| m{12mm} | m{32mm} | m{23mm}| }
\hline
CLUTRR 1.3 & [Elizabeth] bought to dress for his father [Leopold] [Orson] took his wife [Abbie] out for dinner. & father("Leopold", "Orson") \\
\hline
CLUTRR 1.3 & [Asa] felt lonely when his wife [Magdalena] was gone to see her mother [Josiephine]. & wife("Asa", "Magdalena") \\
\hline
CLUTRR 1.3 & [Warner]'s father, [Johnny], and grandfather, [Bryant], went hiking during the first weekend of spring. & male("Johnny") male("Bryant") \\
\hline
CLUTRR 1.3 & [Hollie] and [Rosanna], the happy couple, just got married last week. & -- \\
\hline
CLUTRR 1.3 &  [Violet] took her brother [Travis] to the park, but left her sister [Serena] at home. &  brother("Violet", "Travis") \\
\hline
\end{tabular}
\end{center}

\section{ASP Knowledge Modules}\label{app:sec:asp-modules} 

\begin{table*}[!ht]

\begin{center}
\footnotesize
\begin{tabular}{| l | c | c | c | c |}
\hline
\textbf{Task}              & {\textbf{DEC Axioms}} & {\textbf{Action}} &  {\textbf{Location}} &  {\textbf{Family Relation}}   \\
\cline{1-5}  
1: Single supporting fact  & \checkmark & \checkmark &            &          \\ \hline
2: Two supporting facts    & \checkmark & \checkmark &            &          \\ \hline
3: Three supporting facts  & \checkmark & \checkmark &            &          \\ \hline
4: Two arg relations       &            &            & \checkmark &          \\ \hline
5: Three arg relations     &            & \checkmark &            &          \\ \hline
6: Yes/no questions        & \checkmark & \checkmark &            &          \\ \hline
7: Counting                & \checkmark & \checkmark &            &          \\ \hline
8: Lists/sets              & \checkmark & \checkmark &            &          \\ \hline
9 : Simple negation        & \checkmark & \checkmark &            &          \\ \hline
10: Indefinite knowledge   & \checkmark & \checkmark &            &          \\ \hline
11: Basic coreference      & \checkmark & \checkmark &            &          \\ \hline
12: Conjunction            & \checkmark & \checkmark &            &          \\ \hline
13: Compound coreference   & \checkmark & \checkmark &            &          \\ \hline
14: Time reasoning         & \checkmark & \checkmark &            &          \\ \hline
15: Basic deduction        &            &            &            &          \\ \hline
16: Basic induction        &            &            &            &          \\ \hline
17: Positional reasoning   &            &            & \checkmark &          \\ \hline
18: Size reasoning         &            &            &            &          \\ \hline
19: Path finding           & \checkmark & \checkmark & \checkmark &          \\ \hline
20: Agents motivations     &            &            &            &          \\ \hline
StepGame                   &            &            & \checkmark &          \\ \hline
gSCAN                      & \checkmark &  &   \checkmark         &          \\ \hline
CLUTRR                     &            &            &            & \checkmark \\ \hline
Pick\&Place                & \checkmark & \checkmark &            &          \\ \hline

\end{tabular}
\end{center}
\caption{Knowledge modules used for each of the tasks. Note that {\bf DEC} Axioms, \textbf{action}, and \textbf{location} modules are used in at least two datasets. Some domains aren't listed as they are small and domain specific.}
\label{tb:task_domains}
\end{table*}

\subsection{Discrete Event Calculus (DEC) Axioms Module}

\begin{lstlisting}
% (DEC1)
stopped_in(T1,F,T2) :- timepoint(T),
                       timepoint(T1),
                       timepoint(T2),
                       fluent(F),
                       event(E),
                       happens(E,T),
                       T1<T,
                       T<T2,
                       terminates(E,F,T).

% (DEC2)
started_in(T1,F,T2) :- timepoint(T),
                       timepoint(T1),
                       timepoint(T2),
                       fluent(F),
                       event(E),
                       happens(E,T),
                       T1<T,
                       T<T2,
                       initiates(E,F,T).

% (DEC3)
holds_at(F2,T1+T2) :- timepoint(T1),
                      timepoint(T2),
                      fluent(F1),
                      fluent(F2),
                      event(E),
                      happens(E,T1),
                      initiates(E,F1,T1),
                      0<T2,
                      trajectory(F1,T1,F2,T2),
                      not stopped_in(T1,F1,T1+T2).

% (DEC4)
holds_at(F2,T1+T2) :- timepoint(T1),
                      timepoint(T2),
                      fluent(F1),
                      fluent(F2),
                      event(E),
                      happens(E,T1),
                      terminates(E,F1,T1),
                      0<T2,
                      anti_trajectory(F1,T1,F2,T2),
                      not started_in(T1,F1,T1+T2).

initiated(F,T) :- timepoint(T),
                  fluent(F),
                  event(E),
                  happens(E,T),
                  initiates(E,F,T).

terminated(F,T) :- timepoint(T),
                   fluent(F),
                   event(E),
                   happens(E,T),
                   terminates(E,F,T).

released(F,T) :- timepoint(T),
                 fluent(F),
                 event(E),
                 happens(E,T),
                 releases(E,F,T).

% (DEC5)
holds_at(F,T+1) :- timepoint(T),
                   fluent(F),
                   holds_at(F,T),
                   -released_at(F,T+1),
                   not terminated(F,T).

% (DEC6)
-holds_at(F,T+1) :- timepoint(T),
                    fluent(F),
                    -holds_at(F,T),
                    -released_at(F,T+1),
                    not initiated(F,T).

% (DEC7)
released_at(F,T+1) :- timepoint(T),
                      fluent(F),
                      released_at(F,T),
                      not initiated(F,T),
                      not terminated(F,T).

% (DEC8)
-released_at(F,T+1) :- timepoint(T),
                      fluent(F),
                      -released_at(F,T),
                      not released(F,T).

% (DEC9)
holds_at(F,T+1) :- timepoint(T),
                   fluent(F),
                   event(E),
                   happens(E,T),
                   initiates(E,F,T).

% (DEC10)
-holds_at(F,T+1) :- timepoint(T),
                    fluent(F),
                    event(E),
                    happens(E,T),
                    terminates(E,F,T).

% (DEC11)
released_at(F,T+1) :- timepoint(T),
                      fluent(F),
                      event(E),
                      happens(E,T),
                      releases(E,F,T).

% (DEC12)
-released_at(F,T+1) :- timepoint(T),
                       fluent(F),
                       event(E),
                       happens(E,T),
                       initiates(E,F,T).
-released_at(F,T+1) :- timepoint(T),
                       fluent(F),
                       event(E),
                       happens(E,T),
                       terminates(E,F,T).
\end{lstlisting}

\subsection{Action Module}

\begin{lstlisting}
%********************
* common interface
* check: if location(unknown) is needed
*********************%

% what happened in the given story
happens(action(A, pickup, I), T) :- pickup(A, I, T).
happens(action(A, drop, I), T) :- drop(A, I, T).
happens(action(A1, give, A2, I), T) :- give(A1, I, A2, T).
happens(action(A, goto, L), T) :- go(A, L, T).
happens(action(A, goto, L), T) :- isIn(A, L, T).

%********************
* basic atoms
*********************%

direction(east; west; north; south).

agent(A) :- happens(action(A, _, _), _).
agent(A) :- happens(action(A, _, _, _), _).
agent(A) :- happens(action(_, give, A, _), _).
item(I) :- happens(action(_, pickup, I), _).
item(I) :- happens(action(_, drop, I), _).
item(I) :- happens(action(_, give, _, I), _).
location(L) :- happens(action(_, goto, L), _), not direction(L).

%********************
* atoms in DEC_AXIOMS
*********************%

% event/1
event(action(A, pickup, I)) :- agent(A), item(I).
event(action(A, drop, I)) :- agent(A), item(I).
event(action(A1, give, A2, I)) :- agent(A1), agent(A2), item(I), A1 != A2.
event(action(A, goto, L)) :- agent(A), location(L).
event(action(A, goto, D)) :- agent(A), direction(D).
event(action(robot, pick_and_place, Src, Dst)) :- feature(Src, block), location(Dst), Src != Dst.

% timepoint/1
timepoint(T) :- happens(_, T). % the timepoint in story
timepoint(T) :- T=0..N, maxtime(N). % the timepoint for planning without story

% fluent/1
fluent(at(A, L)) :- agent(A), location(L).
fluent(at(I, L)) :- item(I), location(L).
fluent(carry(A, I)) :- agent(A), item(I).
fluent(on(B, L)) :- feature(B, block), location(L), B!=L.

% -released_at/2
%   1. -released_at(F, T) means commonsense law of inertia (CLI) can be applied to fluent F at T
%   2. CLI is also applied to this literal itself
-released_at(F, 0) :- fluent(F).

% holds_at/2
% initial states of fluents -- only location of items needs to be guessed
{holds_at(at(I, L), 0): location(L)} = 1 :- item(I).
holds_at(on(B, L), 0) :- on(B, L, 0).

% happens/2
% for each timepoint, at most 1 event happens; and it happens as fewer as possible
% {happens(E, T): event(E)}1 :- timepoint(T). % this rule would slow down many tasks
:~ happens(E, T). [1@0, E, T]

% every action should have some effect
%:- happens(E,T), not initiates(E,_,T).

% precondition on actions -- pickup
:- happens(action(A, pickup, I), T), holds_at(at(A, L1), T), holds_at(at(I, L2), T), L1 != L2.

% initiates/3 and terminates/3

% effect of actions -- pickup
initiates(action(A, pickup, I), carry(A, I), T) :- agent(A), item(I), timepoint(T).

% effect of actions -- drop
terminates(action(A, drop, I), carry(A, I), T) :- agent(A), item(I), timepoint(T).

% effect of actions -- give
initiates(action(A1, give, A2, I), carry(A2, I), T) :- agent(A1), agent(A2), item(I), timepoint(T), A1 != A2.
terminates(action(A1, give, A2, I), carry(A1, I), T) :- agent(A1), agent(A2), item(I), timepoint(T), A1 != A2.

% effect of actions -- goto
initiates(action(A, goto, L), at(A, L), T) :- agent(A), location(L), timepoint(T).
initiates(action(A, goto, L), at(I, L), T) :- holds_at(carry(A, I), T), location(L).
initiates(action(A, goto, D), at(A, L2), T) :- 
    agent(A), location(L1), location(L2), timepoint(T), 
    holds_at(at(A, L1), T), is(L2, D, L1).
terminates(action(A, goto, L1), at(A, L2), T) :- agent(A), location(L1), location(L2), timepoint(T), L1 != L2.
terminates(action(A, goto, L1), at(I, L2), T) :- holds_at(carry(A, I), T), location(L1), location(L2), L1 != L2.
terminates(action(A, goto, Direction), at(A, L), T) :-
    happens(action(A, goto, Direction), T),
    holds_at(at(A, L), T), Direction != L.

% effect of actions -- pick_and_place
initiates(action(robot, pick_and_place, Src, Dst), on(Src, Dst), T) :-
    feature(Src, block), location(Dst), Src != Dst, timepoint(T),
    not holds_at(on(_, Src), T), 
    not holds_at(on(_, Dst), T): Dst!="table".

terminates(action(robot, pick_and_place, Src, Dst), on(Src, L), T) :-
    holds_at(on(Src, L), T), location(Dst), Dst != L.
\end{lstlisting}

\subsection{Location Module}\label{appendix:subsec:location_module}
\begin{lstlisting}
% general format translation, which can also be easily done in python script
% (this is not needed if we directly extract the general form in the beginning as in bAbI task4)
is(A, top, B) :- top(A, B).
is(A, top, B) :- up(A, B).
is(A, down, B) :- down(A, B).
is(A, left, B) :- left(A, B).
is(A, right, B) :- right(A, B).
is(A, top_left, B) :- top_left(A, B).
is(A, top_right, B) :- top_right(A, B).
is(A, down_left, B) :- down_left(A, B).
is(A, down_right, B) :- down_right(A, B).
is(A, east, B) :- east(A, B).
is(A, west, B) :- west(A, B).
is(A, south, B) :- south(A, B).
is(A, north, B) :- north(A, B).

% synonyms
synonyms(
    north, northOf; south, southOf; west, westOf; east, eastOf;
    top, northOf; down, southOf; left, westOf; right, eastOf
).
synonyms(A, B) :- synonyms(B, A).
synonyms(A, C) :- synonyms(A, B), synonyms(B, C), A!=C.

% define the offsets of 8 spacial relations
offset(
    overlap,0,0; top,0,1; down,0,-1; left,-1,0; right,1,0; 
    top_left,-1,1; top_right,1,1; down_left,-1,-1; down_right,1,-1
).

% derive the kind of spacial relation from synonyms and offset
is(A, R1, B) :- is(A, R2, B), synonyms(R1, R2).
is(A, R1, B) :- is(B, R2, A), offset(R2,X,Y), offset(R1,-X,-Y).

% derive the location of every object
% the search space of X or Y coordinate is within -100 and 100
% (to avoid infinite loop in clingo when data has error)
nums(-100..100).

location(A, Xa, Ya) :-
    location(B, Xb, Yb), nums(Xa), nums(Ya),
    is(A, Kind, B), offset(Kind, Dx, Dy),
    Xa-Xb=Dx, Ya-Yb=Dy.

location(B, Xb, Yb) :-
    location(A, Xa, Ya), nums(Xb), nums(Yb),
    is_on(A, Kind, B), offset(Kind, Dx, Dy),
    Xa-Xb=Dx, Ya-Yb=Dy.
\end{lstlisting}

\subsection{Family Module}

\begin{lstlisting}
% gender

male(B) :- grandson(A, B).
male(B) :- son(A, B).
male(B) :- nephew(A, B).
male(B) :- brother(A, B).
male(B) :- father(A, B).
male(B) :- uncle(A, B).
male(B) :- grandfather(A, B).

female(B) :- granddaughter(A, B).
female(B) :- daughter(A, B).
female(B) :- niece(A, B).
female(B) :- sister(A, B).
female(B) :- mother(A, B).
female(B) :- aunt(A, B).
female(B) :- grandmother(A, B).

% gender-irrelevant relationships

sibling(A, B) :- siblings(A, B).
sibling(A, B) :- brother(A, B).
sibling(A, B) :- sister(A, B).
sibling(A, B) :- parent(A, C), parent(B, C), A != B.
sibling(A, B) :- sibling(B, A).
sibling(A, B) :- sibling(A, C), sibling(C, B), A != B.
sibling(A, B); sibling_in_law(A, B) :- child(A, C), uncle(C, B).
sibling(A, B); sibling_in_law(A, B) :- child(A, C), aunt(C, B).
sibling_in_law(A, B) :- sibling_in_law(B, A).
:- spouse(A, B), sibling(A, B).
:- spouse(A, B), sibling_in_law(A, B).
:- sibling(A, B), sibling_in_law(A, B).

spouse(A, B) :- wife(A, B).
spouse(A, B) :- husband(A, B).
spouse(A, B) :- spouse(B, A).

parent(A, B) :- father(A, B).
parent(A, B) :- mother(A, B).
parent(A, B) :- parent(A, C), spouse(C, B).
parent(A, B) :- sibling(A, C), parent(C, B).
parent(A, B) :- child(B, A).

child(A, B) :- children(A, B).
child(A, B) :- son(A, B).
child(A, B) :- daughter(A, B).
child(A, B) :- spouse(A, C), child(C, B).
child(A, B) :- child(A, C), sibling(C, B).
child(A, B) :- parent(B, A).

grandparent(A, B) :- grandfather(A, B).
grandparent(A, B) :- grandmother(A, B).
grandparent(A, B) :- parent(A, C), parent(C, B).
grandparent(A, B) :- grandchild(B, A).
grandparent(A, B) :- sibling(A, C), grandparent(C, B).
grandparent(A, B) :- grandparent(A, C), spouse(C, B).

grandchild(A, B) :- grandson(A, B).
grandchild(A, B) :- granddaughter(A, B).
grandchild(A, B) :- grandparent(B, A).

greatgrandparent(A, B) :- grandparent(A, C), parent(C, B).
greatgrandchild(A, B) :- greatgrandparent(B, A).

parent_in_law(A, B) :- spouse(A, C), parent(C, B).
parent(A, B) :- spouse(A, C), parent_in_law(C, B).
parent(A, B); parent_in_law(A, B) :- parent(C, A), grandparent(C, B).
:- parent(A, B), parent(B, A).
:- parent(A, B), parent_in_law(A, B).
child_in_law(A, B) :- parent_in_law(B, A).

% gender-relevant relationships

greatgrandson(A, B) :- greatgrandchild(A, B), male(B).
greatgranddaughter(A, B) :- greatgrandchild(A, B), female(B).

grandson(A, B) :- grandchild(A, B), male(B).
granddaughter(A, B) :- grandchild(A, B), female(B).

son(A, B) :- child(A, B), male(B).
daughter(A, B) :- child(A, B), female(B).
nephew(A, B) :- sibling(A, C), son(C, B).
niece(A, B) :- sibling(A, C), daughter(C, B).

husband(A, B) :- spouse(A, B), male(B).
wife(A, B) :- spouse(A, B), female(B).
brother(A, B) :- sibling(A, B), male(B).
sister(A, B) :- sibling(A, B), female(B).

father(A, B) :- parent(A, B), male(B).
mother(A, B) :- parent(A, B), female(B).
uncle(A, B) :- parent(A, C), brother(C, B).
aunt(A, B) :- parent(A, C), sister(C, B).

grandfather(A, B) :- grandparent(A, B), male(B).
grandmother(A, B) :- grandparent(A, B), female(B).

greatgrandfather(A, B) :- greatgrandparent(A, B), male(B).
greatgrandmother(A, B) :- greatgrandparent(A, B), female(B).

son_in_law(A, B) :- child_in_law(A, B), male(B).
daughter_in_law(A, B) :- child_in_law(A, B), female(B).
father_in_law(A, B) :- parent_in_law(A, B), male(B).
mother_in_law(A, B) :- parent_in_law(A, B), female(B).
\end{lstlisting}

\subsection{Domain Specific Modules}
In this section, we list all domain-specific rules for each task. Some rules serve as an interface to turn the atoms in GPT-3 responses into a general format used in ASP modules. These rules are not necessary and can be removed if we let GPT-3 directly return the general atoms, e.g., ``{\tt query(at(A, where))}'' instead of ``{\tt whereAgent(A)}''. To save the cost for GPT-3 requests, we did not re-produce the experiments using new GPT-3 prompts with atoms in general formats.

\subsubsection{bAbI Tasks 1 and 11}
\begin{lstlisting}
%%%% Interface -- these rules can be removed if we let GPT3 return the heads directly
query(at(A, where)) :- whereAgent(A).

% Find where last location of agent is
answer(L) :- query(at(A, where)), holds_at(at(A, L), T), T>=Tx: holds_at(at(A, _), Tx).
\end{lstlisting}

\subsubsection{bAbI Task 2}

\begin{lstlisting}
%%%% Interface -- these rules can be removed if we let GPT3 return the heads directly
query(at(I, where)) :- loc(I).

% Find where last location of object is
answer(L) :- query(at(A, where)), holds_at(at(A, L), T), T>=Tx: holds_at(at(A, _), Tx).
\end{lstlisting}

\subsubsection{bAbI Tasks 3 and 14}
\begin{lstlisting}
% the query before(O, L) is given, asking about the location of O before moving to L
% find all location changes of the queried object
location_change(L1, L2, T) :- before(O, _), holds_at(at(O, L1), T), holds_at(at(O, L2), T+1), L1 != L2.

% find the last location change to queried location
answer(L1) :- before(_, L2), location_change(L1, L2, T), T>=Tx: location_change(_, L2, Tx).
\end{lstlisting}

\subsubsection{bAbI Task 4}

\begin{lstlisting}
answer(A) :- query(what, R1, B), is(A, R1, B).
answer(B) :- query(A, R1, what), is(A, R1, B).
\end{lstlisting}

\subsubsection{bAbI Task 5}
\begin{lstlisting}
candidate(A1, T) :- query(action(who, give, A, I)), happens(action(A1, give, A2, I), T), 
    A2=A: A!=anyone.
candidate(A2, T) :- query(action(A, give, who, I)), happens(action(A1, give, A2, I), T), 
    A1=A: A!=anyone.
candidate(I, T) :- query(action(A1, give, A2, what)), happens(action(A1, give, A2, I), T).
location(unknown).

%%%% Interface -- these rules can be removed if we let GPT-3 return the heads directly
give(A1, A2, I, T) :- gave(A1, I, A2, T).
query(action(A1, give, A2, what)) :- whatWasGiven(A1, A2).
query(action(anyone, give, who, I)) :- received(I).
query(action(A1, give, who, I)) :- whoWasGiven(A1, I).
query(action(who, give, anyone, I)) :- whoGave(I).
query(action(who, give, A2, I)) :- whoGave(I, A2).

answer(A) :- candidate(A, T), Tx<=T: candidate(_, Tx).
\end{lstlisting}

\subsubsection{bAbI Tasks 6 and 9}
\begin{lstlisting}
answer(yes) :- query(at(A, L)), holds_at(at(A, L), T), Tx<=T: holds_at(at(A, _), Tx).
answer(no) :- not answer(yes).

%%%% Interface -- these rules can be removed if we let GPT-3 return the heads directly
query(at(A, L)) :- isIn(A, L).
\end{lstlisting}

\subsubsection{bAbI Task 7}
\begin{lstlisting}
% find all items I that A is carrying at the last moment; then count I
carry(A, I) :- query(carry(A, count)), holds_at(carry(A,I),T), 
    T>Tx: happens(E,Tx).
location(unknown).

%%%% Interface -- these rules can be removed if we let GPT-3 return the heads directly
query(carry(A, count)) :- howMany(A).

answer(N) :- query(carry(A, count)), N=#count{I: carry(A, I)}.
\end{lstlisting}

\subsubsection{bAbI Task 8}
\begin{lstlisting}
%%%% Interface -- these rules can be removed if we let GPT-3 return the heads directly
query(carry(A, what)) :- carrying(A).
location(unknown).

% find all items I that A is carrying at the last moment
answer(I) :- query(carry(A, what)), holds_at(carry(A,I),T), 
    T>Tx: happens(E,Tx).
\end{lstlisting}

\subsubsection{bAbI Task 10}
\begin{lstlisting}
released(F,T) :- fluent(F), timepoint(T).

answer(yes) :- query(at(A, L)), holds_at(at(A, L), T), Tx<=T: holds_at(at(A, _), Tx).

answer(maybe) :- query(at(A, L)), timepoint(T),
    1{isEither(A, L, _, T); isEither(A, _, L, T)},
    Tx<=T: holds_at(at(A, _), Tx);
    Tx<=T: isEither(A, _, _, Tx).

answer(no) :- not answer(yes), not answer(maybe).

%%%% Interface -- these rules may be removed if we let GPT-3 return the heads directly
query(at(A, L)) :- isInQ(A, L).
holds_at(at(A, L), T) :- isIn(A, L, T).
go(A, L, T) :- move(A, L, T).
timepoint(T) :- isIn(_, _, T).
timepoint(T) :- isEither(_, _, _, T).
\end{lstlisting}

\subsubsection{bAbI Tasks 12 and 13}
\begin{lstlisting}
%%%% Interface -- these rules can be removed if we let GPT-3 return the heads directly
query(at(A, where)) :- whereAgent(A).
go(A1, L, T) :- go(A1, A2, L, T).
go(A2, L, T) :- go(A1, A2, L, T).
\end{lstlisting}

\subsubsection{bAbI Task 15}
\begin{lstlisting}
query(afraid(N, what)) :- agent_afraid(N).
\end{lstlisting}

\subsubsection{bAbI Task 16}

\begin{lstlisting}
animal(frog;lion;swan;rhino).
color(green;white;yellow;gray).
isColor(Agent2,Color):- isAnimal(Agent,Animal),isColor(Agent,Color),isAnimal(Agent2,Animal).
answer(Color) :- isColor(Name), isColor(Name,Color).
\end{lstlisting}

\subsubsection{bAbI Task 17}
\begin{lstlisting}
% assume the 2nd queried object is at location (0,0)
location(B, 0, 0) :- query(_, _, B).

% the queried relation R is correct if its offset agrees with the location of A
answer(yes) :- query(A, R, B), offset(R, Dx, Dy), location(A, X, Y),
    X>0: Dx=1; X<0: Dx=-1;
    Y>0: Dy=1; Y<0: Dy=-1.

answer(no) :- not answer(yes).

%%%% Interface -- these rules can be removed if we let GPT-3 return the heads directly
is(A, left, B) :- leftOf(A, B).
is(A, right, B) :- rightOf(A, B).
is(A, top, B) :- above(A, B).
is(A, down, B) :- below(A, B).
query(A, left, B) :- leftOf_nondirect(A, B).
query(A, right, B) :- rightOf_nondirect(A, B).
query(A, top, B) :- above_nondirect(A, B).
query(A, down, B) :- below_nondirect(A, B).
\end{lstlisting}

\subsubsection{bAbI Task 18}
\begin{lstlisting}
smaller(A, B) :- bigger(B, A).
smaller(A, C) :- smaller(A, B), smaller(B, C).
answer(yes) :- query(smaller(A, B)), smaller(A, B).
answer(no) :- not answer(yes).

%%%% Interface -- these rules can be removed if we let GPT-3 return the heads directly
query(smaller(A, B)) :- doesFit(A, B).
query(smaller(A, B)) :- isBigger(B, A).
\end{lstlisting}

\subsubsection{bAbI Task 19}
\begin{lstlisting}
agent(agent).
maxtime(10).
% location
location(L) :- is(L,_,_).
location(L) :- is(_,_,L).

% for each timestep, we take at most 1 action
{happens(action(agent, goto, D), T): direction(D)}1 :- timepoint(T).

% initial location
holds_at(at(agent, L), 0) :- initial_loc(L).

% goal
:- goal(L), not holds_at(at(agent, L), _).

% we aim to achieve the goal as early as possible
:~ goal(L), holds_at(at(agent, L), T). [-T@1, goal]
\end{lstlisting}

\subsubsection{bAbI Task 20}
\begin{lstlisting}
loc(kitchen). loc(bedroom). loc(kitchen). loc(garden).
obj(pajamas). obj(football). obj(milk). obj(apple).

answer(Location) :- query(where, Agent, go), is(Agent, Quality), motivation(Quality,Location), loc(Location).
answer(Quality) :- query(why, Agent, go, Location), is(Agent, Quality), motivation(Quality, Location), loc(Location).
answer(Quality) :- query(why,Agent, get, Obj),is(Agent, Quality), motivation(Quality, Obj), obj(Obj).
answer(Location) :- query(where, Agent, go), is(Agent, Quality), motivation(Quality, Location), loc(Location).
\end{lstlisting}

\subsubsection{StepGame} \label{appendix:subsec:rules:stepgame}
\begin{lstlisting}
% assume the 2nd queried object is at location (0,0)
location(Q2, 0, 0) :- query(_, Q2).

% extract answer relation R such that the offset (Ox,Oy) of R is in the same direction of (X,Y)
answer(R) :- query(Q1, _), location(Q1, X, Y), offset(R, Ox, Oy),
    Ox=-1: X<0; Ox=0: X=0; Ox=1: X>0;
    Oy=-1: Y<0; Oy=0: Y=0; Oy=1: Y>0.
\end{lstlisting}

\subsubsection{gSCAN}
\begin{lstlisting}
%********************
* find the goal
*********************%

% features of objects
feature(O, shape, V) :- shape(O, V).
feature(O, color, V) :- color(O, V).
feature(O, size, V) :- size(O, V).

% feature of destination
feature(destination, V) :- query(walk), queryDesc(V).
feature(destination, V) :- query(push), queryDesc(V).
feature(destination, V) :- query(pull), queryDesc(V).

% find the destination object and location
pos_same(destination, O) :- feature(O,_,_),
    feature(O,_,V): feature(destination, V), feature(_,_,V).

same(destination, O) :- pos_same(destination, O), feature(O, size, V),
    Vx<=V: feature(destination, big), pos_same(destination, Ox), feature(Ox, size, Vx);
    Vx>=V: feature(destination, small), pos_same(destination, Ox), feature(Ox, size, Vx).

goal(at(agent,L)) :- same(destination, O), pos(O,L).


%********************
* basic atoms
*********************%
agent(agent).
item(I) :-  pos(I, L), I!=agent.
location((X,Y)) :- X=0..N-1, Y=0..N-1, gridSize(N).

is((X1,Y1), east, (X2,Y2)) :- location((X1,Y1)), location((X2,Y2)), X1=X2, Y1=Y2+1.
is((X1,Y1), west, (X2,Y2)) :- location((X1,Y1)), location((X2,Y2)), X1=X2, Y1=Y2-1.
is((X1,Y1), north, (X2,Y2)) :- location((X1,Y1)), location((X2,Y2)), X1=X2-1, Y1=Y2.
is((X1,Y1), south, (X2,Y2)) :- location((X1,Y1)), location((X2,Y2)), X1=X2+1, Y1=Y2.

pos_actions(walk; turn_left; turn_right; stay; push; pull).
left_dir(east, north; north, west; west, south; south, east).

%********************
* atoms in DEC_AXIOMS
*********************%

% fluent/1
fluent(dir(A, L)) :- agent(A), direction(L).
fluent(ready(A)) :- agent(A).

% event/1
event(action(Agent, A)) :- agent(Agent), pos_actions(A).

% initial fluent values
holds_at(at(O,L),0) :- pos(O, L).
holds_at(dir(A,D),0) :- dir(A, D).

% for each timestep, we take at most 1 action
{happens(action(agent, A), T): pos_actions(A)}1 :- timepoint(T).

% initial location
holds_at(at(agent, L), 0) :- initial_loc(L).

%%%%%%%%%%%%%%%
% action -- walk (to check simplification)
%%%%%%%%%%%%%%%

% initiates/3
initiates(action(A, walk), at(A, L2), T) :- agent(A), location(L), timepoint(T),
    holds_at(dir(A, D), T),
    holds_at(at(A, L1), T),
    is(L2, D, L1).

% terminates/3
terminates(action(A, walk), at(A, L1), T) :- agent(A), location(L), timepoint(T),
    holds_at(dir(A, D), T),
    holds_at(at(A, L1), T),
    is(L2, D, L1).

% precondition
% we don't walk in a deadend (i.e., the walk will result in no location change)
:- happens(action(agent, walk), T), not initiates(action(agent, walk), _, T).

%%%%%%%%%%%%%%%
% action -- turn_left (to check simplification)
%%%%%%%%%%%%%%%

% initiates/3
initiates(action(A, turn_left), dir(A, D2), T) :- agent(A), timepoint(T),
    holds_at(dir(A, D1), T),
    left_dir(D1, D2).

% terminates/3
terminates(action(A, turn_left), dir(A, D1), T) :- agent(A), timepoint(T),
    holds_at(dir(A, D1), T).

%%%%%%%%%%%%%%%
% action -- turn_right (to check simplification)
%%%%%%%%%%%%%%%

% initiates/3
initiates(action(A, turn_right), dir(A, D2), T) :- agent(A), timepoint(T),
    holds_at(dir(A, D1), T),
    left_dir(D2, D1).

% terminates/3
terminates(action(A, turn_right), dir(A, D), T) :- agent(A), timepoint(T),
    holds_at(dir(A, D), T).

%%%%%%%%%%%%%%%
% action -- push/pull
%%%%%%%%%%%%%%%

% initiates/3 for objects with size <= 2
initiates(action(A, push), at(A, L2), T) :-
    agent(A), holds_at(at(A, L1), T), holds_at(dir(A, D), T),
    same(destination, Target), holds_at(at(Target, L1), T),
    is(L2, D, L1), feature(Target, size, V), V <= 2.

initiates(action(A, push), at(Target, L2), T) :-
    agent(A), holds_at(at(A, L1), T), holds_at(dir(A, D), T),
    same(destination, Target), holds_at(at(Target, L1), T),
    is(L2, D, L1), feature(Target, size, V), V <= 2.

initiates(action(A, pull), at(A, L2), T) :-
    agent(A), holds_at(at(A, L1), T), holds_at(dir(A, D), T),
    same(destination, Target), holds_at(at(Target, L1), T),
    is(L1, D, L2), feature(Target, size, V), V <= 2.

initiates(action(A, pull), at(Target, L2), T) :-
    agent(A), holds_at(at(A, L1), T), holds_at(dir(A, D), T),
    same(destination, Target), holds_at(at(Target, L1), T),
    is(L1, D, L2), feature(Target, size, V), V <= 2.

% terminates/3 for objects with size <= 2
terminates(action(A, push), at(A, L1), T) :-
    agent(A), holds_at(at(A, L1), T), holds_at(dir(A, D), T),
    same(destination, Target), holds_at(at(Target, L1), T),
    is(L2, D, L1), feature(Target, size, V), V <= 2.

terminates(action(A, push), at(Target, L1), T) :-
    agent(A), holds_at(at(A, L1), T), holds_at(dir(A, D), T),
    same(destination, Target), holds_at(at(Target, L1), T),
    is(L2, D, L1), feature(Target, size, V), V <= 2.

terminates(action(A, pull), at(A, L1), T) :-
    agent(A), holds_at(at(A, L1), T), holds_at(dir(A, D), T),
    same(destination, Target), holds_at(at(Target, L1), T),
    is(L1, D, L2), feature(Target, size, V), V <= 2.

terminates(action(A, pull), at(Target, L1), T) :-
    agent(A), holds_at(at(A, L1), T), holds_at(dir(A, D), T),
    same(destination, Target), holds_at(at(Target, L1), T),
    is(L1, D, L2), feature(Target, size, V), V <= 2.

% initiates/3 for objects with size >= 3

initiates(action(A, push), ready(A), T) :-
    agent(A), holds_at(at(A, L1), T), holds_at(dir(A, D), T), not holds_at(ready(A), T),
    same(destination, Target), holds_at(at(Target, L1), T), feature(Target, size, V), V >= 3.

initiates(action(A, push), at(A, L2), T) :-
    agent(A), holds_at(at(A, L1), T), holds_at(dir(A, D), T), holds_at(ready(A), T),
    same(destination, Target), holds_at(at(Target, L1), T), feature(Target, size, V), V >= 3,
    is(L2, D, L1).

initiates(action(A, push), at(Target, L2), T) :-
    agent(A), holds_at(at(A, L1), T), holds_at(dir(A, D), T), holds_at(ready(A), T),
    same(destination, Target), holds_at(at(Target, L1), T), feature(Target, size, V), V >= 3,
    is(L2, D, L1).

initiates(action(A, pull), ready(A), T) :-
    agent(A), holds_at(at(A, L1), T), holds_at(dir(A, D), T), not holds_at(ready(A), T),
    same(destination, Target), holds_at(at(Target, L1), T), feature(Target, size, V), V >= 3.

initiates(action(A, pull), at(A, L2), T) :-
    agent(A), holds_at(at(A, L1), T), holds_at(dir(A, D), T), holds_at(ready(A), T),
    same(destination, Target), holds_at(at(Target, L1), T), feature(Target, size, V), V >= 3,
    is(L1, D, L2).

initiates(action(A, pull), at(Target, L2), T) :-
    agent(A), holds_at(at(A, L1), T), holds_at(dir(A, D), T), holds_at(ready(A), T),
    same(destination, Target), holds_at(at(Target, L1), T), feature(Target, size, V), V >= 3,
    is(L1, D, L2).

% terminates/3 for objects with size >= 3

{ terminates(action(A, push), ready(A), T);
  terminates(action(A, push), at(A, L1), T);
  terminates(action(A, push), at(Target, L1), T) 
}=3 :-
    agent(A), holds_at(at(A, L1), T), holds_at(dir(A, D), T), holds_at(ready(A), T),
    same(destination, Target), holds_at(at(Target, L1), T), feature(Target, size, V), V >= 3,
    is(L2, D, L1).

{ terminates(action(A, pull), ready(A), T);
  terminates(action(A, pull), at(A, L1), T);
  terminates(action(A, pull), at(Target, L1), T) 
}=3 :-
    agent(A), holds_at(at(A, L1), T), holds_at(dir(A, D), T), holds_at(ready(A), T),
    same(destination, Target), holds_at(at(Target, L1), T), feature(Target, size, V), V >= 3,
    is(L1, D, L2).

% precondition

% 1. we don't push/pull in a deadend (i.e., the action will result in no location change)
:- happens(action(agent, push), T), not initiates(action(agent, push), _, T).
:- happens(action(agent, pull), T), not initiates(action(agent, pull), _, T).

% 2. the agent can push/pull only if it's queried
:- happens(action(agent, push), _), not query(push).
:- happens(action(agent, pull), _), not query(pull).

% 2. it's not allowed to have 3 objects (agent + 2 items) in the same cell
% (I use holds_at(_, T) instead of timepoint(T) since the latter doesn't cover the last T+1 timestamp)
%:- holds_at(_, T), location(L), N = #count{O: holds_at(at(O, L), T)}, N>2.

% 3. after push/pull, the agent cannot do a different action in {walk, push, pull}
:- happens(action(agent, A1), T1), happens(action(agent, A2), T2), A1!=A2, T1<T2,
    1{A1=push; A1=pull},
    1{A2=push; A2=pull; A2=walk}.

% 4. the agent cannot change its direction to push/pull after reaching destination
reach_destination(T) :- goal(at(agent,L)), holds_at(at(agent, L), T),
    not reach_destination(Tx): timepoint(Tx), Tx<T.
:- reach_destination(T1), holds_at(dir(agent, D1), T1), 
    holds_at(dir(agent, D2), T2), happens(action(agent, push), T2), T1<T2, D1!=D2.
:- reach_destination(T1), holds_at(dir(agent, D1), T1), 
    holds_at(dir(agent, D2), T2), happens(action(agent, pull), T2), T1<T2, D1!=D2.

%%%%%%%%%%%%%%%
% goal
%%%%%%%%%%%%%%%

% 1. (optional to speed up) we need to reach the destination and as early as possible
:- goal(at(agent,L)), not reach_destination(_).
:~ goal(at(agent, L)), reach_destination(T). [T@10, goal]

% 2. we need to reach the goal and as early as possible
%   a. the direction when reaching goal must align with the direction when reaching destination
%   b. if it's not deadend, there must be something blocking the next push/pull
reach_goal(T) :- 
    agent(A), holds_at(at(A, L1), T), holds_at(dir(A, D), T),
    same(destination, Target), holds_at(at(Target, L1), T),
    reach_destination(Tr), holds_at(dir(A, D), Tr),
    holds_at(at(_, L2), T): query(push), is(L2, D, L1);
    holds_at(at(_, L2), T): query(pull), is(L1, D, L2);
    not reach_goal(Tx): timepoint(Tx), Tx<T.
:- not reach_goal(_).
:~ reach_goal(T). [T@9, goal]

%%%%%%%%%%%%%%%%%%%%%%%%%%%%%%%%%%%%%%%%%%%%%%%%
% additional requirements to achieve the goal
%%%%%%%%%%%%%%%%%%%%%%%%%%%%%%%%%%%%%%%%%%%%%%%%

% the agent cannot move further before reaching destination
:- reach_destination(T), goal(at(agent, (Xg,Yg))),
    holds_at(at(agent, (X1,Y1)), Tx), holds_at(at(agent, (X2,Y2)), Tx+1), Tx<T,
    |X1-Xg| + |Y1-Yg| < |X2-Xg| + |Y2-Yg|.

% by default, walking all the way horizontally first and then vertically
move(horizontally, T) :- happens(action(agent, walk), T), holds_at(dir(agent, D), T), 1{D=east; D=west}.
move(vertically, T) :- happens(action(agent, walk), T), holds_at(dir(agent, D), T), 1{D=south; D=north}.
:- not while(zigzagging), move(horizontally, T1), move(vertically, T2), T1>T2.

% hesitantly: the agent must stay after every action in {walk, push, pull}
:- while(hesitantly), happens(action(agent, A), T),
    1{A=walk; A=push; A=pull},
    not happens(action(agent, stay), T+1).

% cautiously
cautious(T) :- happens(action(agent, turn_left), T), 
    happens(action(agent, turn_right), T+1),
    happens(action(agent, turn_right), T+2),
    happens(action(agent, turn_left), T+3).

% the agent must be cautious before every action in {walk, push, pull}
:- while(cautiously), happens(action(agent, A), T),
    1{A=walk; A=push; A=pull},
    not cautious(T-4).

% spinning
spin(T) :- happens(action(agent, turn_left), T), 
    happens(action(agent, turn_left), T+1),
    happens(action(agent, turn_left), T+2),
    happens(action(agent, turn_left), T+3).

% we always spin at the beginning if there is any action
:- while(spinning), happens(_,_), not spin(0).

% we always spin after every action in {walk, push, pull} except for the last one
:- while(spinning), happens(action(agent, A1), T1), happens(action(agent, A2), T2), T1<T2,
    1{A1=walk; A1=push; A1=pull},
    1{A2=walk; A2=push; A2=pull},
    not spin(T1+1).

% zigzagging
% if horizontal move is needed, the first move must be horizontal
:- while(zigzagging), move(horizontally, _), move(D, Tmin), D!=horizontally,
    Tmin<=Tx: move(_,Tx).
% if a different kind of move D2 is after D1, D2 must be followed directly
:- while(zigzagging), move(D1, T1), move(D2, T2), D1!=D2, T1<T2,
    not move(D2, T1+2).
\end{lstlisting}

\subsubsection{Pick\&Place}
\begin{lstlisting}
%%%%%
% Set up the environment
%%%%%

% Define the number of grippers for the robot
#const grippers=1.

% Define the maximum number of steps to consider
{maxtime(M): M=0..10} = 1.
:~ maxtime(M). [M]

%%%%%
% Extract the features for all items in the intial and goal states
% we assume these items form the complete set of items in this example
%%%%%

feature(I, F) :- on(I,_), F=@gen_feature(I).
feature(I, F) :- on(I,_,0), F=@gen_feature(I).
feature(I, F) :- on(_,I), I!="table", F=@gen_feature(I).
feature(I, F) :- on(_,I,0), I!="table", F=@gen_feature(I).

% Define all locations
location("table").
location(L) :- feature(L, block).
location(L) :- feature(L, bowl).

%********************
* atoms in DEC_AXIOMS
*********************%

% happens/2
{happens(E,T): event(E)}grippers :- timepoint(T).

% **** constraints ****

% the goal must be achieved in the end
:- maxtime(M), on(A, B), not holds_at(on(A,B), M+1).

% At any time T, for each block/bowl, there cannot be 2 items directly on it
:- timepoint(T), feature(L, _), 2{holds_at(on(I, L), T): feature(I,_)}.

% if there are bowls on the table, a block can only be on a block or a bowl;
:- feature(_,bowl), feature(I,block), holds_at(on(I,L),_), {feature(L, block); feature(L, bowl)} = 0.

% there cannot be more than max_height-1 blocks stacked on a block
up(A,B,T) :- holds_at(on(A, B), T).
up(A,C,T) :- up(A,B,T), up(B,C,T).
:- timepoint(T), feature(L, block), #count{I: up(I,L,T)} >= max_height.
\end{lstlisting}

\section{Dataset Errors}\label{app:sec:dataset-errors} 

This section enumerates the errors in the datasets we found. 

\subsection{bAbI}\label{appendix:Dataset_errors:bAbI}

In task 5, the dataset has two errors with regard to the labels.

\noindent\textbf{{Error \#1}}. In the following example, the answer is ambiguous since Bill gives Mary both the football and the apple. 
\begin{lstlisting}
CONTEXT:
Mary journeyed to the kitchen.
Mary went to the bedroom.
Mary moved to the bathroom.
Mary grabbed the football there.
Mary moved to the garden.
Mary dropped the football.
Fred went back to the kitchen.
Jeff went back to the office.
Jeff went to the bathroom.
Bill took the apple there.
Mary picked up the milk there.
Mary picked up the football there.
Bill went back to the kitchen.
Bill went back to the hallway.
Fred journeyed to the office.
Bill discarded the apple.
Mary journeyed to the kitchen.
Fred journeyed to the garden.
Mary went to the hallway.
Mary gave the football to Bill.
Bill passed the football to Mary.
Bill took the apple there.
Bill gave the apple to Mary.
Jeff travelled to the kitchen.

QUERY:
What did Bill give to Mary?

PREDICTION:
apple

Answer:
football


\end{lstlisting}

\noindent{\textbf{Error \#2.}}
In the following example, the answer is ambiguous since Fred gives Bill both the milk and the apple. 
\begin{lstlisting}
CONTEXT:
Mary journeyed to the bathroom.
Mary moved to the hallway.
Mary went to the kitchen.
Bill went back to the bedroom.
Bill grabbed the apple there.
Fred went back to the garden.
Mary went to the garden.
Fred took the milk there.
Jeff moved to the hallway.
Bill dropped the apple there.
Fred handed the milk to Mary.
Mary handed the milk to Fred.
Fred went back to the bedroom.
Fred passed the milk to Bill.
Fred took the apple there.
Fred gave the apple to Bill.
Jeff went to the kitchen.
Bill dropped the milk.

QUERY:
What did Fred give to Bill?

PREDICTION:
apple

Answer:
milk
\end{lstlisting}

\subsection{CLUTRR}\label{appendix:data_errors:clutrr}
We detected 16 data errors in the CLUTRR 1.3 dataset using our method. These errors can be grouped into the following 4 categories.
\begin{itemize}
\item 5 data instances are due to incorrect relation graphs. For example, one relation graph contains the main part ``A-son-B-daughter-C-aunt-D'' and a noise (supporting) relation ``B-spouse-D''. However, if B and D are couples, then C should have mother D instead of aunt D.
\item 9 data instances have a correct relation graph (e.g., A-son-B-grandmother-C-brother-D with a noise supporting relation B-mother-A) but the noise relation is translated into a sentence with a wrong person name (e.g., "D has mother A" instead of "B has mother A").
\item 1 data instance has a correct relation graph and story, but has a wrong label (i.e., the label should be mother\_in\_law instead of mother).
\item 1 data instance has a correct relation graph and story, but the query cannot be answered due to the ambiguity of a sentence.  It uses "A has grandsons B and C" to represent brother(B, C), while B and C may have different parents.
\end{itemize}

\end{document}